\documentclass[10pt,twocolumn,letterpaper]{article}
\usepackage[pagenumbers]{cvpr} %

\definecolor{cvprblue}{rgb}{0.21,0.49,0.74}
\usepackage[pagebackref,breaklinks,colorlinks,allcolors=cvprblue]{hyperref}
\usepackage[frozencache,cachedir=minted-cache]{minted}
\usepackage{array}
\usepackage{algorithm}
\usepackage[noend]{algorithmic}

\usepackage{multirow}
\usepackage{colortbl}
\newcommand{\grayline}{\arrayrulecolor[HTML]{BFBFBF}\hline\arrayrulecolor{black}}
\def\paperID{9439} %
\def\confName{CVPR}
\def\confYear{2025}

\title{Multi-layer Radial Basis Function Networks for Out-of-distribution Detection}

\author{Amol Khanna$^*$\\
Booz Allen Hamilton\\
Boston, MA\\
{\tt\small Khanna\_Amol@bah.com}
\and
Chenyi Ling$^*$\\
Booz Allen Hamilton\\
Columbia, MD\\
{\tt\small Ling\_Chenyi@bah.com}
\and 
Derek Everett\\
Booz Allen Hamilton\\
Columbia, MD\\
{\tt\small Everett\_Derek@bah.com}
\and 
Edward Raff\\
Booz Allen Hamilton\\
Syracuse, NY\\
{\tt\small Raff\_Edward@bah.com}
\and 
Nathan Inkawhich\\
Air Force Research Laboratory\\
Rome, NY\\
{\tt\small Nathan.Inkawhich@us.af.mil}
}

\begin{document}
\maketitle
\def\thefootnote{*}\footnotetext{These authors contributed equally to this work. \\ Approved for Public Release; Distribution Unlimited. PA Number:
AFRL-2024-6733.}\def\thefootnote{\arabic{footnote}}
\begin{abstract}
Existing methods for out-of-distribution (OOD) detection use various techniques to produce a score, separate from classification, that determines how ``OOD'' an input is. Our insight is that OOD detection can be simplified by using a neural network architecture which can effectively merge classification and OOD detection into a single step. Radial basis function networks (RBFNs) inherently link classification confidence and OOD detection; however, these networks have lost popularity due to the difficult of training them in a multi-layer fashion. In this work, we develop a multi-layer radial basis function network (MLRBFN) which can be easily trained. To ensure that these networks are also effective for OOD detection, we develop a novel depression mechanism. We apply MLRBFNs as standalone classifiers and as heads on top of pretrained feature extractors, and find that they are competitive with commonly used methods for OOD detection. Our MLRBFN architecture demonstrates a promising new direction for OOD detection methods. 

\end{abstract}    
\section{Introduction}

Common OOD detection methods fall into three categories. There are \emph{post-hoc} methods which try to extract signals from already-trained models to separate in-distribution (ID) and OOD data. For example, by using the softmax output as a proxy for uncertainty, the maximum softmax probability method classifies inputs with lower maximum softmax probabilities as OOD \cite{hendrycks2016baseline}. \emph{Modified training} methods engineer loss functions or employ generative models to encourage networks to learn representations of ID data which are robust to variations. During inference, these methods compare the representations of an input to known representations, and if these are sufficiently different, the input is marked as OOD \cite{ming2022exploit,du2022vos,tao2023non}. Finally, if OOD data is available at training time, \emph{outlier exposure} methods teach networks to explicitly produce distinguishable representations for ID and OOD inputs \cite{hendrycks2018deep}. 

\begin{figure}[t]
    \centering
    \begin{subfigure}{0.75\linewidth}
        \centering
        \includegraphics[width=\linewidth]{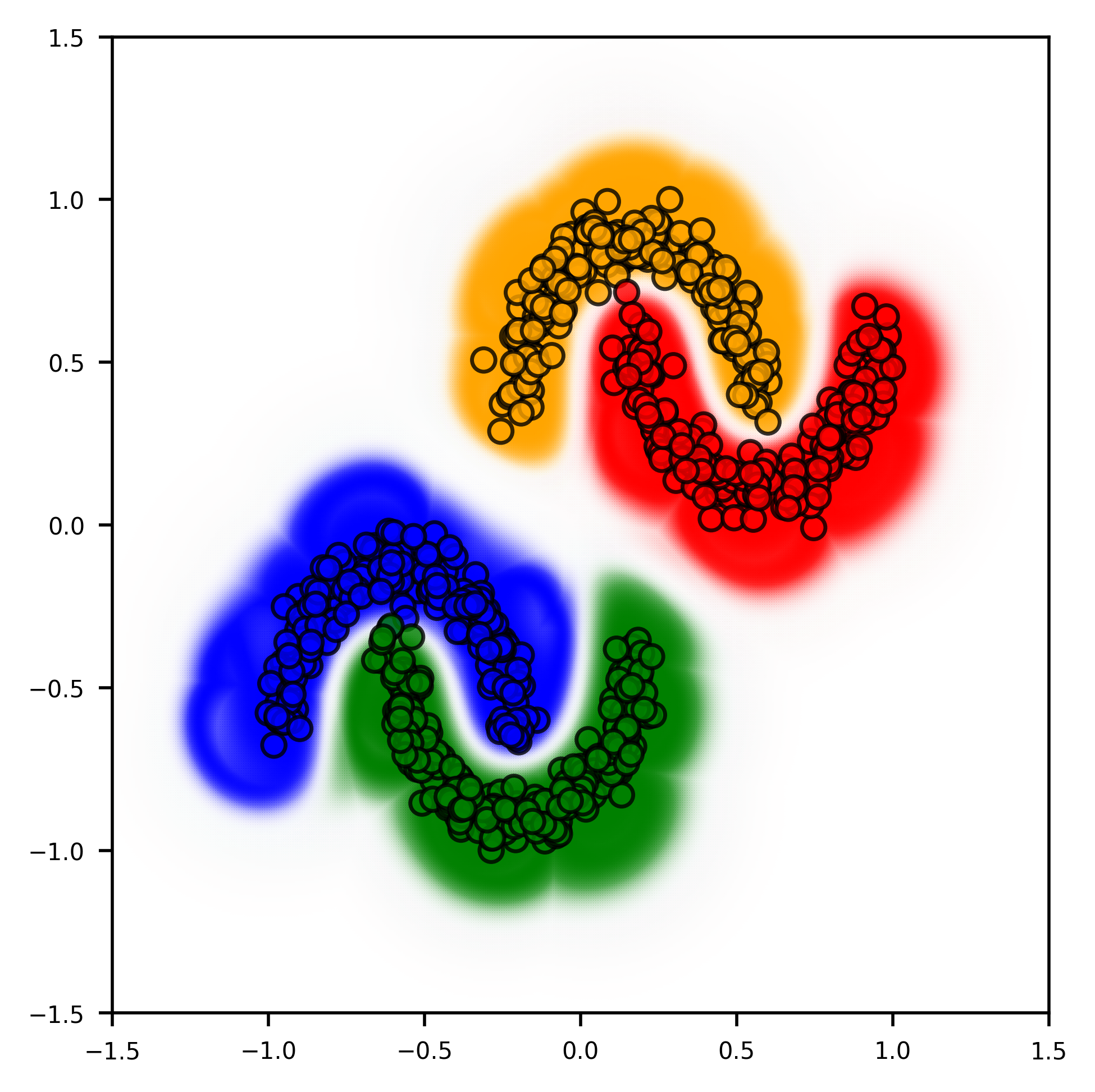}
    \end{subfigure}
    \caption{MLRBFNs are capable of correct classification with high confidence on the 4-class moons dataset. Away from the training data manifold, the network is not confident.}
    \label{fig:teaser}
\end{figure}

\begin{figure}[t]
    \centering
    \begin{subfigure}{0.49\linewidth}
        \centering
        \includegraphics[width=\linewidth]{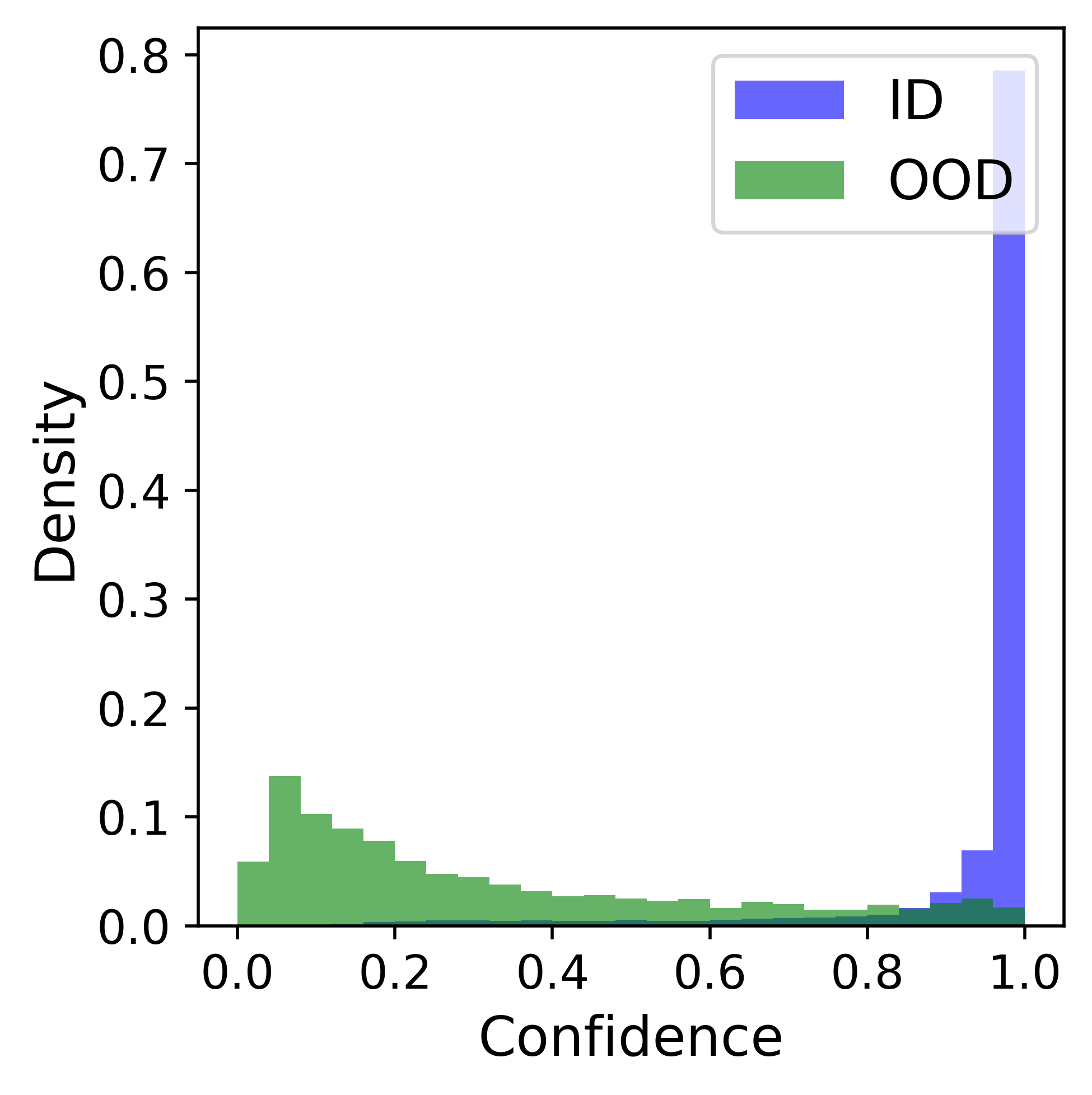}
        \caption{Near-OOD}
    \end{subfigure}
    \hfill
    \begin{subfigure}{0.49\linewidth}
        \centering
        \includegraphics[width=\linewidth]{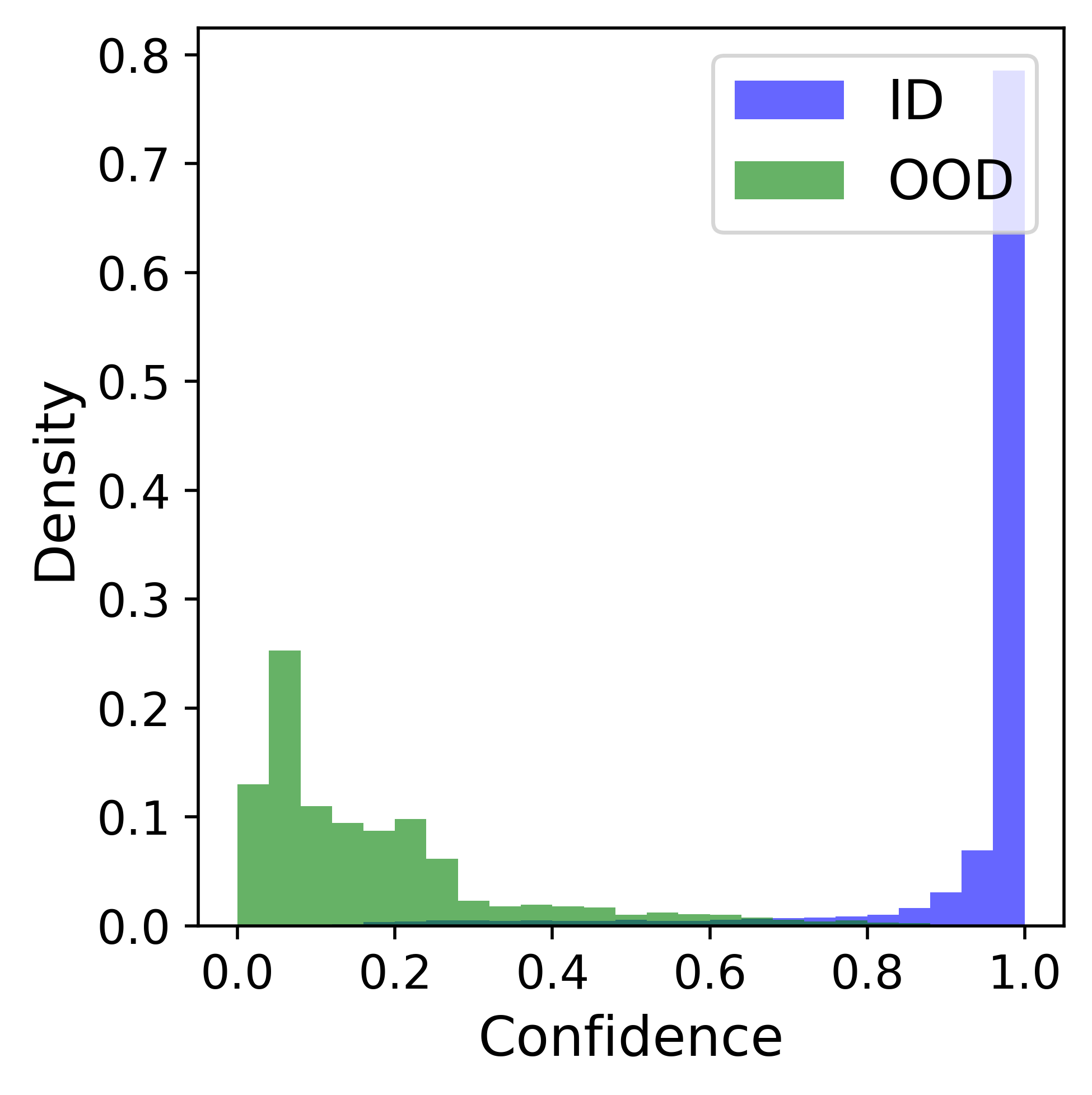}
        \caption{Far-OOD}
    \end{subfigure}
    \caption{Confidence of ID and OOD datasets when training an MLRBFN. In both subfigures, CIFAR-10 is the ID dataset, and CLIP ViT-B/16 is used to extract features. (a) shows the frequency of scores for the near-OOD dataset Tiny ImageNet, and (b) shows the scores for the far-OOD dataset Texture. The MLRBFN demonstrates strong separability between the ID and both OOD datasets.
    }
    \label{fig:cifar10}
\end{figure}

In all three cases, the mechanism for OOD detection is distinct from the classifier itself; the OOD detector will have its own true- and false-positive rates. In this work, we explore a strategy to make OOD detection an intrinsic component of the classification process and neural network architecture, such that sufficiently OOD inputs will receive no classification because they are OOD, and OOD inputs are detected by noting that the maximal class score is not sufficiently high. This idea is illustrated in \autoref{fig:teaser}, where the range of reasonable predictions only exists near the data manifold. \autoref{fig:cifar10} demonstrates that this approach works on the more challenging CIFAR10 dataset.

To achieve this goal, we develop a new approach to training multi-layer Radial Basis Function Networks (RBFNs), which will be applied as a head to a pre-trained feature extractor. RBFNs are one layer networks which classify datapoints based on their distances to different learned centroids. However, extending this architecture to multi-layer RBFNs (MLRBFNs) has been notoriously difficult to train effectively while retaining RBFNs' inherent capacity for OOD detection. We identify a key difficulty as the \emph{$\boldsymbol{0}$-mapped class}: where one class is represented by the network using the zero vector, which intrinsically represents all ``other'' items. We introduce a new technique called the \emph{depression mechanism} that mitigates this issue and allows learning MLRBFNs that perform well at OOD detection using standard backpropagation, which has been historically elusive.

The structure of this paper is as follows. In \autoref{sec:relatedwork}, we discuss the challenges of OOD detection with modern Deep Neural Network (DNN) layers and introduce RBFNs. Next, \autoref{sec:methods} details our primary contribution - methods to resolve the challenges of OOD detection using MLRBFNs. We demonstrate the performance of our MLRBFN architecture in \autoref{sec:results}. Finally, we conclude in \autoref{sec:conclusion}. We believe our MLRBFN architecture is a significant step towards neural networks which can inherently detect OOD datapoints.

\section{Related Work}
\label{sec:relatedwork}

\subsection{Challenges for OOD Detection on Modern DNNs}

It is well known that modern DNNs are miscalibrated and produce high confidence predictions far away from the training data manifold \cite{guo2017calibration,wang2023calibration,bartlett2008classification,cortes2009learning}. Many works have attempted to rectify miscalibration using noisy gradient updates, Bayesian layers, or data augmentations, but fewer have characterized the source of the overconfidence in DNN architectures \cite{ferianc2024making,rahimi2020post,lakshminarayanan2017simple,liu2020simple}. \cite{hein2019relu} identifies that DNNs have \emph{changing and confident} predictions far away from the training data manifold\footnote{See Figure 1 of \cite{hein2019relu} for an excellent visualization.}, and attributes this behavior to the linear classification boundaries of modern (ReLU) DNNs. Indeed, the authors suggest that RBFNs could mitigate these issues. 

In addition to modifying DNN architectures, loss functions can be used to promote calibration. Inspired by classic statistical literature on scoring rules, recent works have focused on defining and studying proper loss functions which can incentivize calibration \cite{blasiok2024does}.\footnote{Facing an OOD input on a 4-class classification problem, a DNN trained with a proper loss function should produce predictions of $\begin{bmatrix} \frac{1}{4} & \frac{1}{4} & \frac{1}{4} & \frac{1}{4} \end{bmatrix}^\top$.} However, common proper loss functions for multiclass classification such as cross-entropy or negative log-likelihood encode a bias that every input should be classified as one of $C$ classes in a $C$-class problem, and this bias gets transferred to the learned DNN after training. In contrast, inspired by hypothesis testing, traditional statistical scoring rules were designed to produce calibration on binary classification problems \cite{gneiting2007strictly,liu2020simple}. Splitting $C$-class classification problems into $C$ binary classification problems and employing a scoring rule, we expect a calibrated DNN facing an OOD input to produce predictions of the zero vector ($\mathbf{0}$), indicating zero confidence on each class \cite{cheng2024towards}. In this work, we seek to produce effective confidence estimates by splitting the classification task into $C$ binary classification problems and using the binary cross-entropy loss, which is a statistical scoring rule \cite{waggoner2017}. By doing this, we allow the network to output $\mathbf{0}$, indicating that an input is dissimilar from all training classes. We take this step to avoid biasing our model towards bucketing all inputs into one of $C$ classes. However, previous works have incorporated this reasoning with only incremental improvements \cite{bitterwolf2022revisiting}. Our work is unique since we use RBF activations which can naturally promote OOD detection. 

\subsection{RBF Networks}

RBF Networks are artificial neural networks that employ radial basis functions for activation and were popular in the 1990s \cite{lowe1988multivariable}. These networks compute the distances from an input to a set of learned centroids and use a linear layer to classify inputs based on these distances. Extending the RBFN architecture to the multi-layer setting was elusive and prevented further interest. 

The structure of an RBFN typically consists of an RBF activation function followed by a linear projection. Specifically, we write the forward pass of a datapoint through an RBFN as 

\begin{align*}
    s_c(\mathbf{x}) &= \exp \left( -\beta_c^+ \lVert \mathbf{x} - \mathbf{c}_c \rVert_k^k \right) \\ 
    \phi(\mathbf{x}) &= \sum_{c = 1} ^ {N} \mathbf{a}_c s_c(\mathbf{x}),
\end{align*}
where the RBFN has $N$ centroids. Note that $\mathbf{x} \in \mathbb{R}^d$, $\mathbf{c}_c \in \mathbb{R}^d$ and $\mathbf{a}_c \in \mathbb{R}^o$, where $d$ is the input dimension and $o$ is the output dimension of the network. $\beta_c^+$ is typically constrained to be greater than 0, as is represented by the ${}^+$, and any $k \geq 1$ norm can be used for the distance computation.\footnote{Positivity constraints for $\beta_c^+$ can be achieved by using $\beta_c^+ = \texttt{softplus}(\beta_c)$.} For clarity, $\mathbf{c}_c$ is the \textit{position} of the $c^\text{th}$ centroid, $s_c(\mathbf{x})$ is the \textit{weight} with respect to the $c^\text{th}$ centroid, and $\mathbf{a}_c$ is the \textit{projection} associated with the $c^\text{th}$ centroid. We also refer to $\beta_c^+$ as the \textit{inverse-width} of the $c^\text{th}$ centroid due to its similarity to precision parameters in the Gaussian distribution. Note that $s_c$ produces a localized response, which is what allows the network to produce $\boldsymbol{0}$ as a response to dissimilar content \cite{powell1977restart}.

Training an RBFN involves two primary tasks: determining the centroids and inverse-widths of the RBFN and adjusting the projections for output. Traditionally, the centroids of the RBFN are chosen based on clustering techniques such as $k$-means, which partition the data into regions where each basis function can effectively represent local patterns. Inverse-widths are chosen by determining the distance of a centroid to its respective datapoints, and the projections are calculated using linear or logistic regression \cite{schwenker2001three}. 

If centroids and inverse-widths are trained appropriately, it is clear that RBFNs can inherently detect OOD inputs. OOD inputs will be far from every centroid, and assuming inverse-widths are fit tightly to each cluster, each $\beta_c^+$ will be as large as possible. This means $s_c(\mathbf{x}) = \exp \left( -\beta_c^+ \lVert \mathbf{x} - \mathbf{c}_c \rVert_k^k \right) \approx 0$ for every $c$, so $\phi(\mathbf{x}) \approx \mathbf{0}$ for OOD inputs.

Despite the rise of DNNs, the inherent robustness of RBFNs to OOD inputs makes them an attractive option for reconsideration in modern machine learning. Unlike DNNs, which can be overconfident on OOD inputs, RBFNs naturally exhibit caution due to their reliance on distance metrics \cite{hein2019relu}. However, RBFNs have limited scalability to complex problems due to their shallow architecture. Our MLRBFN architecture enables deep RBFNs which allows for improved scalability to complex problems. 

\subsection{Distance-Awareness and Multi-Layer \\RBF Networks Unfit for OOD Detection}

Prior works have considered the relationship between OOD detection and distance-awareness. \cite{liu2020simple} identifies that \emph{distance-awareness}, or methods being  aware of the distance between an input and previous training examples, is critical for OOD detection and uncertainty quantification. The authors note that DNNs do not maintain distance-awareness in their representations and show that spectral normalization can improve their behavior in this aspect. However, their approach is dissimilar from ours since they do not use RBFNs. Additionally, their method is still prone to feature collapse in DNN linear layers, which can reduce effectiveness of OOD detection. A related method called Deterministic Uncertainty Quantification (DUQ) employs a single RBF activation in the last layer of a DNN \cite{van2020uncertainty}. This method suffers from the same issues as above: the representations produced by DNN linear layers are prone to feature collapse and do not maintain distance-awareness, and training the RBF layer relies on more heuristic approaches compared to our approach that allows backpropogation through a multi-layer RBF network.

Next, there are a few methods which claim to use ``deep'' RBFNs but actually only use one RBF layer along with standard convolutional neural networks or multi-layer perceptrons \cite{zadeh2018deep,chen2018deep}. Another work employs elementwise RBF activations \cite{hryniowski2019deeplabnet}. However, due to a large number of centroids, these networks overfit rapidly. They also require very custom initialization strategies and use kernels designed to work far away from centroids. None of these works focus on uncertainty quantification, miscalibration, or OOD detection.  

The most relevant work to ours is \cite{wurzberger2024learning}, which details the architecture for deep RBF transformations. The authors describe a network with structure 
\begin{align}
\label{eq:pub-mlrbfn}
    \phi_{\ell}(\mathbf{x}) = \begin{bmatrix} \exp \left( -\beta_{1(\ell)}^+ \lVert \phi_{\ell - 1}(\mathbf{x}) - \mathbf{c}_{1(\ell)} \rVert_k^k \right) \\ \vdots \\ \exp \left( -\beta_{N(\ell)}^+ \lVert \phi_{\ell - 1}(\mathbf{x}) - \mathbf{c}_{N(\ell)} \rVert_k^k \right) \end{bmatrix}
\end{align}
with a linear projection layer $\phi_{L}(\mathbf{x}) = \mathbf{W}\phi_{L - 1}(\mathbf{x})$ at the end of the network. Note that $\ell = 1, \dots, L - 1, L$ are the layers of the network and $\phi_0(\mathbf{x}) = \mathbf{x}$. In this formulation, $c(\ell)$ indexes the $c^{\text{th}}$ centroid in the $\ell^{\text{th}}$ layer. They demonstrate that this network is trainable with backpropogation on the MNIST and CIFAR10 datasets. However, they do not consider OOD detection, and in our testing, this architecture does not retain the OOD detection capacity of single-layer RBFNs due to \fbox{$\mathbf{0}$-mapped classes:}

\begin{figure}[t]
    \centering
    \includegraphics[width=0.75\linewidth]{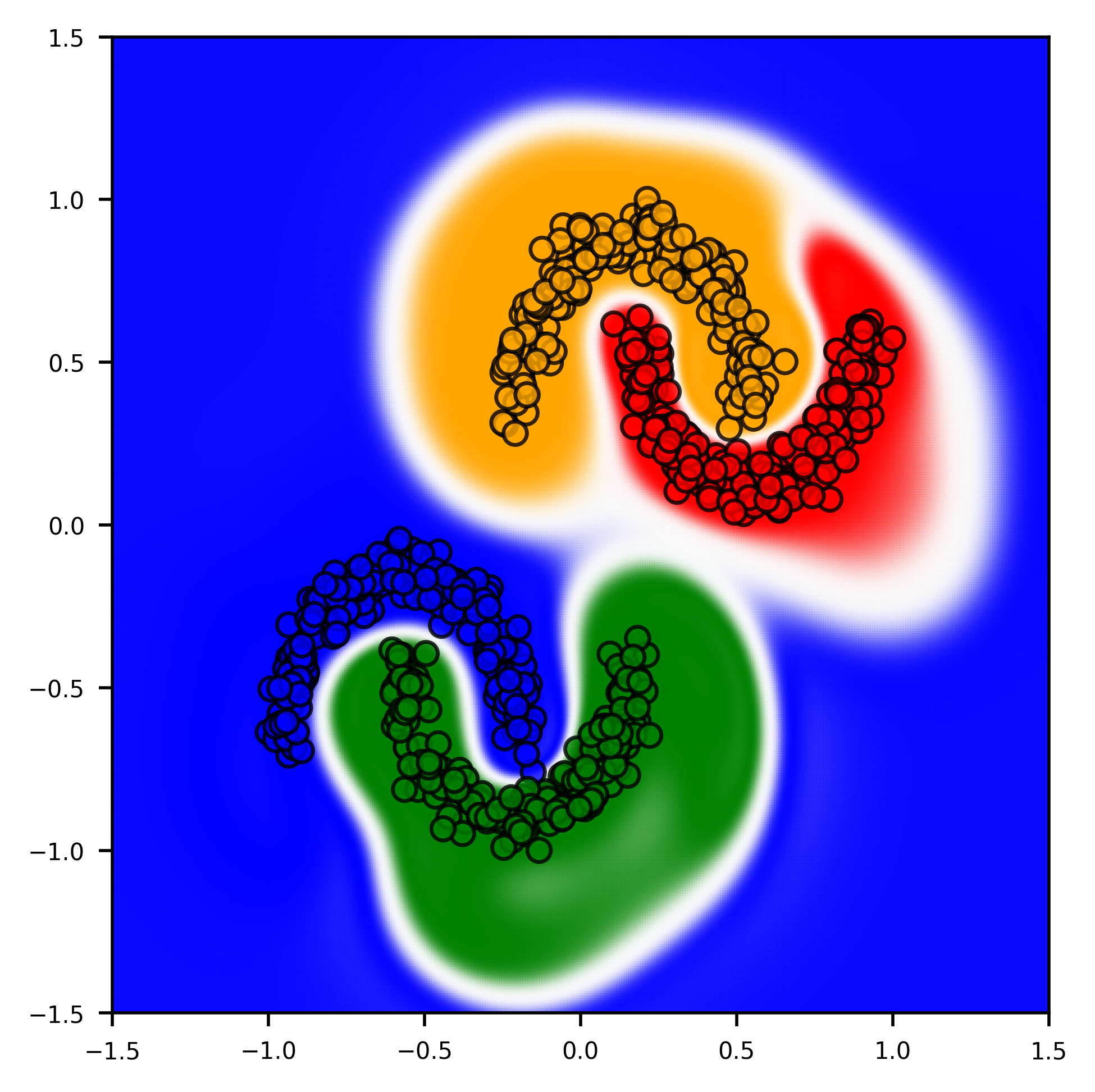}
    \caption{An MLRBFN trained on the 4-class moons dataset without the depression mechanism. Employing class-wise binary cross-entropy teaches the network to only predict one class with high confidence near each moon, but the network has very wide confidence regions near each moon. It is also very confident that inputs far away from all moons should be predicted as the blue class. Compare this to \autoref{fig:teaser}, in which the depression mechanism produces confidence regions are restricted to around the data manifold.}
    \label{fig:moons-nodepression}
\end{figure}

When employing \autoref{eq:pub-mlrbfn} to learn a $C$-class problem, we would like the MLRBFN to learn centroids near the manifold of training data at each layer. This would allow the MLRBFN to produce high-confidence predictions near the training manifold but approximately $\mathbf{0}$ away from this manifold, enabling OOD detection. Unfortunately, this is not what happens. Training with backpropogation, the MLRBFN learns to use its centroids to only model $C - 1$ classes, having no centroids near one class in the input and subsequent layers. This ``left out'' class is mapped to $\mathbf{0}$ after every layer until the penultimate layer, when the MLRBFN puts a centroid at $\mathbf{0}$ in the last layer to map this class to its correct prediction. While this produces an accurate solution, it is ineffective for OOD detection because points in the input domain far from the training manifold are also mapped to $\mathbf{0}$ in the first $L - 2$ layers, and are then predicted as the left out class with high confidence in the $(L - 1)^{\text{th}}$ or $L^\text{th}$ layers. We call this behavior $\mathbf{0}$-mapped classes, and a visual example can be seen in \autoref{fig:moons-nodepression}. One of our main contributions in this paper is the \emph{depression mechanism}, which addresses this issue and is described in \autoref{sec:methods}. 

\section{Methods}
\label{sec:methods}

Inspired to develop effective MLRBFNs for OOD detection, we modify the architecture described in \autoref{eq:pub-mlrbfn} to this end. Note that we found training to be much more stable when employing a linear projection after every RBF transformation, and thus in the following subsections we develop an MLRBFN based off the following structure: 
\begin{align}
\label{eq:naive-mlrbfn}
    s_{c(\ell)}(\mathbf{x}) &= \exp \left( -\beta_{c(\ell)}^+ \lVert \phi_{\ell - 1}(\mathbf{x}) - \mathbf{c}_{c(\ell)} \rVert_k^k \right) \nonumber \\ 
    \phi_\ell(\mathbf{x}) &= \sum_{c = 1} ^ {N} \mathbf{a}_{c(\ell)} s_{c(\ell)}(\mathbf{x}),
\end{align}
This formulation is still prone to $\mathbf{0}$-mapped classes, and we discuss our modifications in this section. Note that you can find the code for our RBFN layers in \autoref{sec:code}. 

\subsection{Decoupling Predictions and Confidence}

As described in \autoref{sec:relatedwork}, DNNs trained with cross-entropy or negative log-likelihood are implicitly biased towards predicting a learned class on every input, regardless of whether it is OOD. Our MLRBFN training methods address these issues in two ways: First, for each training datapoint, we convert a $C$-class classification problem to $C$ binary classification problems, and employ the binary-cross entropy loss on each of these $C$ problems. This decouples the predictions of each class and explicitly teaches the network to predict that datapoints may \emph{not} be of certain classes. Next, we have the network output probabilities for each class in its final layer.\footnote{Our code actually outputs \emph{log-probabilities} in the final layer, since this enhances numerical stability with the binary-cross entropy loss.} To do this, in the final layer we do not use projections and instead set 
\begin{equation}
\label{eq:da}
    \phi_{c(L)}(\mathbf{x}) = \exp \left( -\beta_{c(L)}^+ \lVert \phi_{L - 1}(\mathbf{x}) - \mathbf{c}_{c(L)} \rVert_k^k \right)
\end{equation} and use $C$ centroids in this layer, one for each class. Since this expression is between $0$ and $1$, each of $\phi_{1(L)}(\mathbf{x}), \phi_{2(L)}(\mathbf{x}), \dots, \phi_{C(L)}(\mathbf{x})$ can be interpreted as the network's confidence that an input is in the $1^{\text{st}}, 2^{\text{nd}}, \dots, C^{\text{th}}$ class. 

\subsection{Preventing Zero-mapped Classes}

To address the issue of $\mathbf{0}$-mapped classes detailed in \autoref{sec:relatedwork}, we develop a \emph{depression mechanism}. The following paragraphs describe the three steps we took to ensure the depression mechanism is effective. The final formulation is shown in \autoref{eq:depression}. 

The depression mechanism must ensure that intermediate representations of an MLRBFN which have been mapped to $\mathbf{0}$ cannot be mapped to a high-confidence at the output of the MLRBFN. In other words, $\mathbf{0}$ is a ``black hole:'' once a datapoint is mapped to $\mathbf{0}$, it will continue being mapped to $\mathbf{0}$. Mathematically, we formulate the depression mechanism as 
\begin{align}
\label{eq:naive-depression}
    s_{c(\ell)}(\mathbf{x}) &= \exp\left( -\beta_{c(\ell)}^+ \lVert \phi_{\ell - 1}(\mathbf{x}) - \mathbf{c}_{c(\ell)} \rVert_k^k \right) \nonumber \\ 
    n_{c(\ell)}(\mathbf{x}) &= s_{c(\ell)}(\mathbf{x}) \times \min \left(\text{dep}_{\ell - 1}(\mathbf{x}) \times \text{rec}, 1.0 \right) \nonumber \\ 
    \text{dep}_{\ell}(\mathbf{x}) &= \max_{c(\ell)} s_{c(\ell)}(\mathbf{x}) \times \min \left(\text{dep}_{\ell - 1}(\mathbf{x}) \times \text{rec}, 1.0 \right) \nonumber \\ 
    \phi_\ell(\mathbf{x}) &= \sum_{c = 1}^{N} \mathbf{a}_{c(\ell)} n_{c(\ell)}(\mathbf{x})
\end{align}
where dep standards for depression and rec stands for recovery. We return $\phi_\ell(\mathbf{x})$ and $\text{dep}_{\ell}(\mathbf{x})$ after each layer, and initialize $\text{dep}_0$ to 1. We set rec to a small constant greater than 1, for example 1.05. 

To provide intuition for \autoref{eq:naive-depression}, note the following. When an intermediate representation $\phi_{\ell - 1}(\mathbf{x})$ is far away from all centroids in layer $\ell$, $\max_{c(\ell)} s_{c(\ell)}(\mathbf{x}) \approx 0$. This means that $\text{dep}_\ell(\mathbf{x}) \approx 0$. Thus, in the next layer, $n_{c(\ell + 1)}(\mathbf{x}) \approx 0$, meaning that $\phi_{\ell + 1}(\mathbf{x}) \approx 0$. Finally, once $\text{dep}_\ell(\mathbf{x}) \approx 0$, this low value of the depression gets propogated since the depression update is recursive. 

We develop the depression mechanism so MLRBFNs learn centroids near intermediate representations of training datapoints of all classes to produce accurate and confident solutions. Concurrently, for OOD inputs, the depression mechanism should force $\phi_{\ell}(\mathbf{x})$ toward $\mathbf{0}$ since the representations of OOD inputs are far from learned centroids. This, however, is not what occurs. Instead, the network circumvents the intended behavior by learning at least one centroid with very small inverse-width in each layer. For these centroids, $s_{c(\ell)}(\mathbf{x}) = \exp\left( -\beta_{c(\ell)}^+ \lVert \phi_{\ell - 1}(\mathbf{x}) - \mathbf{c}_{c(\ell)} \rVert_k^k \right) \approx 1$ even when $\lVert \phi_{\ell - 1}(\mathbf{x}) - \mathbf{c}_{c(\ell)} \rVert_k^k$ is very large, meaning that in each layer, the depression mechanism will not yield an effective penalty. To resolve this issue, we must penalize small inverse-widths. 

To do this, we take inspiration from the Gaussian distribution. The Gaussian PDF integrates to a finite value because it multiplies an inverse-width factor with its exponential term. This inverse-width factor ensures that the peak of the Gaussian becomes smaller as the Gaussian grows wider. Using this reasoning, we set $s_{c(\ell)}(\mathbf{x}) = \beta_{c(\ell)}^+ \exp\left( -\beta_{c(\ell)}^+ \lVert \phi_{\ell - 1}(\mathbf{x}) - \mathbf{c}_{c(\ell)} \rVert_k^k \right)$, incorporating an inverse-width factor into the weight calculation. By doing this, we ensure that centroids with very small inverse-widths will both be penalized by the depression calculation and will have minimal impact on the projection calculation, eliciting the desired behavior from the MLRBFN. 

When testing this formulation, we find that the MLRBFN training becomes unstable. To stabilize it, we instead employed $s_{c(\ell)}(\mathbf{x}) = \frac{\beta_{c(\ell)}^+}{\beta_{{c(\ell)}_{\text{init}}}^+} \exp\left( -\beta_{c(\ell)}^+ \lVert \phi_{\ell - 1}(\mathbf{x}) - \mathbf{c}_{c(\ell)} \rVert_k^k \right)$, which replaces the inverse-width factor by a ratio of the trained inverse-width to the inverse-width set after initialization but prior to training, as described in \autoref{sec:initialization}. This ensures that during initial training iterations, the network focuses on learning its centroid and projection parameters rather than being burdened by the inverse-width factor. 

For clarity, we rewrite the depression mechanism with the inverse-width penalty here: 
\begin{align}
\label{eq:depression}
    s_{c(\ell)}(\mathbf{x}) &= \frac{\beta_{c(\ell)}^+}{\beta_{{c(\ell)}_{\text{init}}}^+} \exp\left( -\beta_{c(\ell)}^+ \lVert \phi_{\ell - 1}(\mathbf{x}) - \mathbf{c}_{c(\ell)} \rVert_k^k \right) \nonumber \\ 
    n_{c(\ell)}(\mathbf{x}) &= s_{c(\ell)}(\mathbf{x}) \times \min \left(\text{dep}_{\ell - 1}(\mathbf{x}) \times \text{rec}, 1.0 \right) \nonumber \\ 
    \text{dep}_{\ell}(\mathbf{x}) &= \max_{c(\ell)} s_{c(\ell)}(\mathbf{x}) \times \min \left(\text{dep}_{\ell - 1}(\mathbf{x}) \times \text{rec}, 1.0 \right) \nonumber \\ 
    \phi_\ell(\mathbf{x}) &= \sum_{c = 1}^{N} \mathbf{a}_{c(\ell)} n_{c(\ell)}(\mathbf{x})
\end{align}
To incorporate depression into the final predictions, we use 
\begin{align}
\label{eq:depression-finallayer}
    s_{c(L)}(\mathbf{x}) &= \exp \left( -\beta_{c(L)}^+ \lVert \phi_{L - 1}(\mathbf{x}) - \mathbf{c}_{c(L)} \rVert_k^k \right) \nonumber \\ 
    \phi_{c(L)}(\mathbf{x}) &= s_{c(L)}(\mathbf{x}) \times \min \left(\text{dep}_{L - 1}(\mathbf{x}) \times \text{rec}, 1.0 \right)
\end{align}
for the final layer. The results we present use this formulation during training and inference.

\subsection{Quicker Training with Clustering}

Finally, on the first forward pass through the MLRBFN, we set the centroids and inverse-widths of every layer with $k$-means clustering. Details can be found in \autoref{sec:initialization}. 

\begin{algorithm}[t]
\caption{Overview of MLRBFN Training}
\label{alg:mlrbfn-training}
\begin{algorithmic}[1]
\small
\STATE Initialize centroids and inverse-widths with one batch
\FOR{$e = 1$ \TO epochs}
    \FOR{$(\mathbf{X}, \mathbf{y})$ \textbf{in} batches}
        \STATE $\phi_0(\mathbf{X}) = \mathbf{X},\  \text{dep}_0 (\mathbf{X}) = \mathbf{1}$
        \FOR{$\ell = 1$ \TO $L - 1$}
            \STATE $\phi_{\ell}(\mathbf{X}),\  \text{dep}_{\ell}(\mathbf{X}) = \text{Eq. }\ref{eq:depression} \left( \phi_{\ell - 1}(\mathbf{X}),\  \text{dep}_{\ell - 1}(\mathbf{X}) \right)$
        \ENDFOR
        \STATE $\phi_{1(L)}, \dots \phi_{C(L)} = \text{Eq. }\ref{eq:depression-finallayer}\left(\phi_{L - 1}(\mathbf{X}), \text{dep}_{L - 1}(\mathbf{X}) \right)$
        \STATE $l = 0$
        \FOR{$c = 1$ \TO $C$}
            \STATE $l = l + \text{Bin. Cross-Ent.} \left( \phi_{c(L)}, \text{One-Hot}_c(\mathbf{y}) \right)$
        \ENDFOR
        \STATE Backpropogate on $l$ and update
    \ENDFOR
\ENDFOR
\end{algorithmic}
\end{algorithm}

An overview of MLRBFN training can be found in \autoref{alg:mlrbfn-training}. Note that $\text{One-Hot}_c$ is a one-hot encoding function which assigns $1$ to class $c$ and $0$ to all other classes. 
\section{Results}
\label{sec:results}

We tested our MLRBFN formulation in 2 ways. First, we used MLRBFNs as a standalone ``deep'' model. Second, we used MLRBFNs as classification ``heads'' which take as input features extracted from pretrained embedding models. 

We first demonstrate standalone MLRBFNs on the MNIST dataset, and illustrate why MLRBFNs are promising. We also discuss some of their current limitations: a lack of convolutions and vanishing gradients with increasing layer depth. These two challenges cause MLRBFNs to not be independently ready for larger datasets and are the most pressing areas for future research. 

For this reason, we use MLRBFNs as a head applied to pretrained feature extractors, as is common in modern OOD detection literature. In doing so, we find that MLRBFNs are highly competitive with relevant benchmarks. Specifically, on every benchmark task, directly using MLRBFN outputs as scores for OOD detection outperforms using the maximum softmax probability from standard DNNs for OOD detection. This indicates that MLRBFNs are inherently better capable of detecting OOD inputs than naive DNNs. 

\subsection{Standalone MLRBFNs}

\begin{figure}
    \centering
    \includegraphics[width=0.75\linewidth]{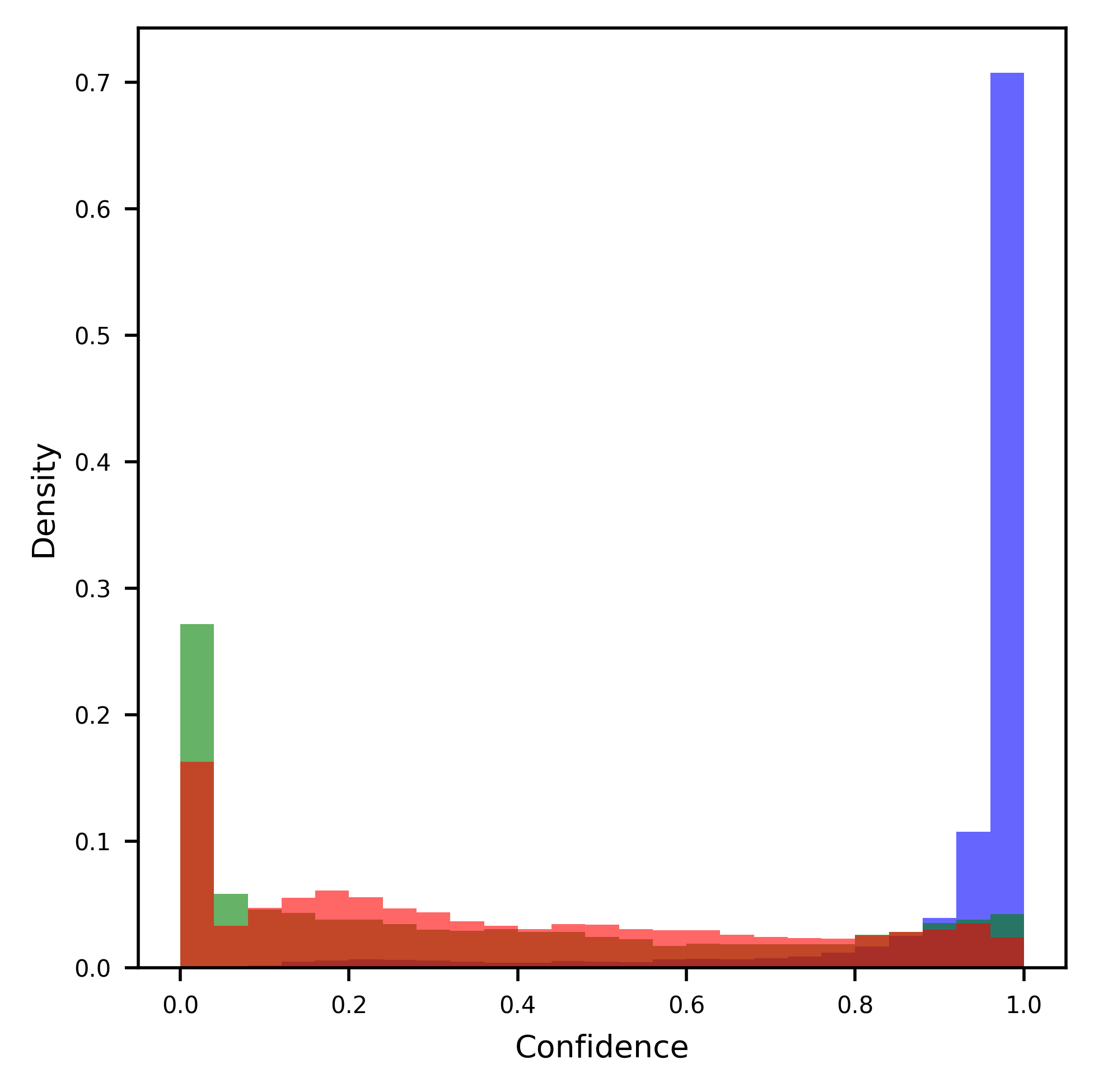}
    \caption{Confidences of an MLRBFN trained on the MNIST dataset. MNIST testing data confidence is \textcolor[HTML]{6666F6}{blue}, while OOD datasets FashionMNIST and KMNIST are in \textcolor[HTML]{78B06E}{green} and \textcolor[HTML]{ED706B}{red}, respectively. It is clear that MLRBFNs are capable of distinguishing between ID and OOD inputs throu}
    \label{fig:mnist}
\end{figure}

The first benchmark we tested with our MLRBFN architecture was the 4-moons dataset. Results for the moons dataset are discussed in \autoref{sec:moons} due to space limitations. 

Next, we used MLRBFNs for classification on the MNIST dataset. We trained a 4-layer MLRBFN on the normalized and flattened MNIST dataset with a batch size of 256 \cite{deng2012mnist}. The first three layers of the MLRBFN had $N = 200$ centroids while the last had $N = C = 10$ centroids. A projection dimension of $o = 100$ was used for the first three layers. $k$ was set to 2 for all layers. The Adam optimizer was used with a learning rate of $10^{-3}$, and the network was trained for 200 epochs. After training, the network had $>99.5\%$ training accuracy and $>97.5\%$ testing accuracy.

\autoref{fig:mnist} demonstrates the results of this experiment. As expected, the ID dataset has high confidence, and the FashionMNIST and KMNIST OOD datasets have lower confidence, with many OOD datapoints having confidences around 0 \cite{xiao2017/online,clanuwat2018deep}. We compute an AUROC of 0.95 for both datasets, and a FPR@95 of 0.32 and 0.35 for FashionMNIST and KMNIST, respectively. 

The results for MNIST are competitive with the best baselines but do not surpass them \cite{yang2022openood}. We believe that this is because our network is quite shallow and cannot learn effective features for images due to lack of convolutions. Although we would like to train a deeper network, we find that our MLRBFN formulation is prone to gradient vanishing due to the depression mechanism.\footnote{Each layer in the depression mechanism involves multiplication of a value between $0$ and $1$. This is the same reason that sigmoid and tanh activations are prone to gradient vanishing.} We recognize that identifying an MLRBFN formulation which is not prone to gradient vanishing is an important direction for future research. However, we believe that our current MLRBFN formulation is still valuable since it is competitive with state-of-the-art OOD detectors when applied to features extracted from foundation models. 

\subsection{MLRBFNs as Classification Heads}

Since the depression mechanism is prone to gradient vanishing, we do not use very deep MLRBFN architectures. To achieve effective classification and OOD detection with MLRBFNs on complex datasets, we take advantage of recent advances in foundation models. Research shows that many deep learning tasks can be accomplished without training models from scratch. Instead, foundation models can act as powerful feature extractors and only task-specific classification heads need to be learned on top of these features. In this spirit, we employ pretrained foundation models to extract features from images which are then used as inputs to MLRBFNs. 

\subsubsection{Feature Extractors and Training Procedure}

The performance of common OOD detection algorithms is strongly influenced by the feature extractor architecture \cite{kim2024comparison}. The OpenOOD v1.5 review employs ResNet18 and ResNet50 backbones for the CIFAR10/CIFAR100/ImageNet200 and ImageNet1K datasets, respectively \cite{zhang2023openood}. However, we believe that these architectures will not effectively highlight the potential of MLRBFNs. 

When employing a pretrained feature extractor prior to using MLRBFNs, the feature extractor must produce separable representations for the features of different classes. Additionally, the features of OOD inputs must be distinguishable from those of ID classes. From the field of transfer learning, we know that the first condition often holds - pretrained feature extractors are commonly used for downstream classification tasks \cite{weiss2016survey}. However, on feature extractors trained on image classification tasks, the second condition rarely holds due to neural and feature collapse, a phenomenon where neural networks will produce feature representations and outputs similar to those learned from their training data regardless of whether an input is ID or OOD \cite{kothapalli2022neural,papyan2020prevalence,zhu2021geometric,laurent2023feature,van2021feature}. Indeed, some studies suggest that the classification task and its associated loss functions encourage DNNs to learn the simplest features to separate the ID classes, but these featues do not promote the separability of ID and OOD representations \cite{parker2023neural,kothapalli2022neural,zhou2022all}. For this reason, we believe that MLRBFNs will have the best OOD detection performance when using features extracted from general-purpose foundation models. These models are likely less prone to feature collapse because they are both trained on significantly more data than classification models and they do not employ classification loss functions during training.\footnote{We remark that this is a hypothesis - we cannot find a sufficiently comprehensive study of feature collapse on foundation models to support or refute this theory.} To demonstrate the importance of feature extractors, we choose to use ResNet18 and ResNet50 backbones trained on ImageNet along with OpenAI's CLIP (ViT-B/16) and Meta's DINOv2 (ViT-B/14) as frozen feature extractors \cite{he2016deep,radford2021learning,oquab2023dinov2}.

To train the MLRBFNs, we first generated datasets containing features extracted from these models. We conducted training experiments on both small and large in-distribution (ID) datasets, including CIFAR-10, CIFAR-100, ImageNet-200, and ImageNet-1K. These datasets were taken from the OpenOOD v1.5 library which processes datasets to remove overlapping samples from ID and OOD datasets \cite{zhang2023openood}. The library also provides code and datsets for separate evaluation on near-OOD problems (eg. CIFAR100 for a network trained on CIFAR10) or far-OOD problems (eg. MNIST for a network trained on CIFAR10). We used OpenOOD's near- and far-OOD datasets.

For training on CIFAR10, CIFAR100, and ImageNet200, we used an MLRBFN with $4$ layers and a batch size of $128$. Each layer of the network was set with $N = 50$ centroids and a projection dimension of $o = 100$. We used an Adam optimizer with learning rate $10^{-3}$ accompanied by a ReduceLROnPlateau scheduler. The scheduler reduced the learning rate by a factor of $0.5$ if the validation loss did not improve for $50$ consecutive epochs. Training was performed for $500$ epochs. For training on ImageNet1K, we used an MLRBFN with $5$ layers. The batch size, number of centroids, projection dimensions, and optimization was the same as above, but we trained for $1000$ epochs.

Visualization of MLRBFN performance are shown in \autoref{fig:cifar10}. AUPR Out and FPR@95 results are deferred to \autoref{sec:add_results} due to constrained space. Below we describe results on the AUROC metric when comparing to different benchmarks. 

\subsubsection{Comparing to OpenOOD v1.5}

\begin{table*}[!ht]
\caption{Comparing the performance of MLRBFNs to published benchmarks and existing methods in the OpenOOD v1.5 study. In this table, we use AUROC as the metric for OOD detection. Whenever possible, we report the average number and the corresponding standard deviation obtained from 3 training runs. In almost all cases, MLRBFNs using the ResNet18 (for CIFAR10/CIFAR100/ImageNet200) and ResNet50 (for ImageNet1K) backbones are competitive with the average performance of methods from the OpenOOD v1.5 study. MLRBFNs using DINOv2 and CLIP as feature extracts generally outperform the best performing methods from OpenOOD v1.5.
}
\label{tab:openood_results}
\centering
{\renewcommand\baselinestretch{1.3}\selectfont\resizebox{\textwidth}{!}{
    \begin{tabular}{l|>{\centering\arraybackslash}p{1.6cm}>{\centering\arraybackslash}p{1.5cm}>{\centering\arraybackslash}p{1.5cm}|>{\centering\arraybackslash}p{1.6cm}>{\centering\arraybackslash}p{1.5cm}>{\centering\arraybackslash}p{1.5cm}|>{\centering\arraybackslash}p{1.6cm}>{\centering\arraybackslash}p{1.5cm}>{\centering\arraybackslash}p{1.5cm}|>{\centering\arraybackslash}p{1.6cm}>{\centering\arraybackslash}p{1.5cm}>{\centering\arraybackslash}p{1.5cm}} 
    \toprule
     & \multicolumn{3}{c|}{\textbf{CIFAR10}} & \multicolumn{3}{c|}{\textbf{CIFAR100}} & \multicolumn{3}{c|}{\textbf{ImageNet200}} & \multicolumn{3}{c}{\textbf{ImageNet1K}} \\
    
     & Near-OOD & Far-OOD & ID Acc. & Near-OOD & Far-OOD & ID Acc. & Near-OOD & Far-OOD & ID Acc. & Near-OOD & Far-OOD & ID Acc. \\
    \midrule

\multicolumn{7}{l}{\textbf{- Multi-layer Radial Basis Function Networks}} \vspace{.1cm} \\

ResNet & 88.88\textsubscript{(\textpm 0.26)} & 90.01\textsubscript{(\textpm 0.58)} & 94.63\textsubscript{(\textpm 0.19)} & 77.30\textsubscript{(\textpm 0.49)} & 80.51\textsubscript{(\textpm 1.18)} & 69.64\textsubscript{(\textpm 0.30)} & 81.02\textsubscript{\hphantom{(\textpm 0.00)}} & 89.10\textsubscript{\hphantom{(\textpm 0.00)}} & 84.50\textsubscript{\hphantom{(\textpm 0.00)}} & 74.30\textsubscript{\hphantom{(\textpm 0.00)}} & 86.60\textsubscript{\hphantom{(\textpm 0.00)}} & 70.72\textsubscript{\hphantom{(\textpm 0.00)}} \\

DINOv2 & 96.01\textsubscript{(\textpm 0.34)} & 97.60\textsubscript{(\textpm 0.64)} & 97.79\textsubscript{(\textpm 0.04)} & 89.18\textsubscript{(\textpm 0.73)} & 93.18\textsubscript{(\textpm 1.17)} & 85.25\textsubscript{(\textpm 0.32)} & 90.24\textsubscript{(\textpm 0.22)} & 96.82\textsubscript{(\textpm 0.52)} & 94.61\textsubscript{(\textpm 0.37)} & 78.84\textsubscript{(\textpm 0.25)} & 92.57\textsubscript{(\textpm 0.21)} & 78.62\textsubscript{(\textpm 0.12)} \\

CLIP & 95.75\textsubscript{(\textpm 0.35)} & 98.46\textsubscript{(\textpm 0.20)} & 95.78\textsubscript{(\textpm 0.16)} & 86.53\textsubscript{(\textpm 0.84)} & 95.96\textsubscript{(\textpm 1.35)} & 76.04\textsubscript{(\textpm 0.17)} & 85.05\textsubscript{(\textpm 0.42)} & 94.78\textsubscript{(\textpm 0.42)} & 90.97\textsubscript{(\textpm 0.07)} & 72.03\textsubscript{(\textpm 0.51)} & 88.50\textsubscript{(\textpm 0.32)} & 72.16\textsubscript{(\textpm 0.11)} \\

\multicolumn{7}{l}{\textbf{- OpenOOD v1.5 Benchmarks}} \vspace{.1cm} \\
Best Post-hoc & 90.64\textsubscript{(\textpm 0.20)} & 93.48\textsubscript{(\textpm 0.24)} & 95.06\textsubscript{(\textpm 0.30)} & 81.05\textsubscript{(\textpm 0.07)} & 82.92\textsubscript{(\textpm 0.42)} & 77.25\textsubscript{(\textpm 0.10)} & 83.69\textsubscript{(\textpm 0.04)} & 93.90\textsubscript{(\textpm 0.27)} & 86.37\textsubscript{(\textpm 0.08)} & 78.17\textsubscript{\hphantom{(\textpm 0.00)}} & 95.74\textsubscript{\hphantom{(\textpm 0.00)}} & 76.18\textsubscript{\hphantom{(\textpm 0.00)}} \\

Best Mod. Training & 92.68\textsubscript{(\textpm 0.27)} & 96.74\textsubscript{(\textpm 0.06)} & 95.35\textsubscript{(\textpm 0.52)} & 80.93\textsubscript{(\textpm 0.29)} & 88.40\textsubscript{(\textpm 0.13)} & 77.20\textsubscript{(\textpm 0.10)} & 82.66\textsubscript{(\textpm 0.15)} & 94.49\textsubscript{(\textpm 0.07)} & 86.37\textsubscript{(\textpm 0.16)} & 76.52\textsubscript{\hphantom{(\textpm 0.00)}} & 92.18\textsubscript{\hphantom{(\textpm 0.00)}} & 76.55\textsubscript{\hphantom{(\textpm 0.00)}} \\

Best Outlier Exp. & 94.82\textsubscript{(\textpm 0.21)} & 96.00\textsubscript{(\textpm 0.13)} & 94.95\textsubscript{(\textpm 0.04)} & 88.30\textsubscript{(\textpm 0.10)} & 81.41\textsubscript{(\textpm 1.49)} & 76.84\textsubscript{(\textpm 0.42)} & 84.84\textsubscript{(\textpm 0.16)} & 89.02\textsubscript{(\textpm 0.18)} & 86.12\textsubscript{(\textpm 0.07)} & \textcolor{gray}{N/A}\textsubscript{\hphantom{(\textpm 0.00)}} & \textcolor{gray}{N/A}\textsubscript{\hphantom{(\textpm 0.00)}} & \textcolor{gray}{N/A}\textsubscript{\hphantom{(\textpm 0.00)}} \\

Avg. Post-hoc & 79.18\textsubscript{\hphantom{(\textpm 0.00)}} & 83.24\textsubscript{\hphantom{(\textpm 0.00)}} & 95.06\textsubscript{\hphantom{(\textpm 0.00)}} & 73.16\textsubscript{\hphantom{(\textpm 0.00)}} & 76.43\textsubscript{\hphantom{(\textpm 0.00)}} & 77.25\textsubscript{\hphantom{(\textpm 0.00)}} & 75.88\textsubscript{\hphantom{(\textpm 0.00)}} & 84.02\textsubscript{\hphantom{(\textpm 0.00)}} & 86.37\textsubscript{\hphantom{(\textpm 0.00)}} & 71.11\textsubscript{\hphantom{(\textpm 0.00)}} & 85.36\textsubscript{\hphantom{(\textpm 0.00)}} & 76.18\textsubscript{\hphantom{(\textpm 0.00)}} \\ 

Avg. Mod. Training & 88.06\textsubscript{\hphantom{(\textpm 0.00)}} & 91.91\textsubscript{\hphantom{(\textpm 0.00)}} & 94.15\textsubscript{\hphantom{(\textpm 0.00)}} & 76.28\textsubscript{\hphantom{(\textpm 0.00)}} & 78.88\textsubscript{\hphantom{(\textpm 0.00)}} & 73.74\textsubscript{\hphantom{(\textpm 0.00)}} & 79.44\textsubscript{\hphantom{(\textpm 0.00)}} & 90.47\textsubscript{\hphantom{(\textpm 0.00)}} & 85.52\textsubscript{\hphantom{(\textpm 0.00)}} & 72.96\textsubscript{\hphantom{(\textpm 0.00)}} & 87.35\textsubscript{\hphantom{(\textpm 0.00)}} & 75.36\textsubscript{\hphantom{(\textpm 0.00)}} \\

Avg. Outlier Exp. & 91.12\textsubscript{\hphantom{(\textpm 0.00)}} & 93.25\textsubscript{\hphantom{(\textpm 0.00)}} & 94.12\textsubscript{\hphantom{(\textpm 0.00)}} & 81.09\textsubscript{\hphantom{(\textpm 0.00)}} & 78.03\textsubscript{\hphantom{(\textpm 0.00)}} & 74.84\textsubscript{\hphantom{(\textpm 0.00)}} & 81.35\textsubscript{\hphantom{(\textpm 0.00)}} & 87.08\textsubscript{\hphantom{(\textpm 0.00)}} & 81.44\textsubscript{\hphantom{(\textpm 0.00)}} & \textcolor{gray}{N/A}\textsubscript{\hphantom{(\textpm 0.00)}} & \textcolor{gray}{N/A}\textsubscript{\hphantom{(\textpm 0.00)}} & \textcolor{gray}{N/A}\textsubscript{\hphantom{(\textpm 0.00)}} \\

\bottomrule
    \end{tabular}}\par}
\end{table*}

We begin by comparing MLRBFNs to the methods reviewed in the OpenOOD v1.5 study. Results are shown in \autoref{tab:openood_results}. In the table, we provide two benchmarks - first, we list the best observed performance for each column of the methods tested in the OpenOOD v1.5 paper, separated by method type. Second, we provide the average performance of each type of detector. This is because different methods achieved the best performance for each column in the results listed in the OpenOODv1.5 paper. By using the average performance of established OOD detection methods, we can estimate how well a single OOD detection method performs across all columns and also compare this to MLRBFNs. 

\autoref{tab:openood_results} shows that MLRBFNs using ResNet feature extractors generally outperform average OOD detector performance and are competitive with the best OOD detectors on most datasets. This is despite MLRBFNs using a frozen ResNet feature extractors while the OpenOOD study finetuned these backbones for each OOD detection method. This result lends faith to the design of MLRBFNs to inherently detect OOD inputs. 

We also include results for the DINOv2 and CLIP feature extractors in \autoref{tab:openood_results}. These results consistently outperform the best performing OOD detectors from OpenOOD v1.5 using the ResNet18 and ResNet50 backbones. This demonstrates that feature extractors play a significant role in OOD detection, with more powerful extractors being able to separate ID and OOD inputs better.  

\subsubsection{Comparing to Postprocessors Using Foundation Models' Features}

\begin{table*}[!ht]
\caption{Comparing the performance of MLRBFNs to published benchmarks and existing methods. In this table, we use AUROC as the metric for OOD detection. Whenever applicable, we report the average number and the corresponding standard deviation obtained from 3 training runs. The table demonstrates that MLRBFNs outperform MSP in all cases (emphasized in \textcolor{red}{red}), meaning that the outputs of MLRBFNs are better suited for OOD detection than standard DNNs. This result supports our goal of building a neural network architecture which is better able to inherently detect OOD inputs than modern DNNs. 
}
\label{tab:foundation_results}
\centering
{\renewcommand\baselinestretch{1.3}\selectfont\resizebox{\textwidth}{!}{
    \begin{tabular}{ll|>{\centering\arraybackslash}p{1.6cm}>{\centering\arraybackslash}p{1.5cm}>{\centering\arraybackslash}p{1.5cm}|>{\centering\arraybackslash}p{1.6cm}>{\centering\arraybackslash}p{1.5cm}>{\centering\arraybackslash}p{1.5cm}|>{\centering\arraybackslash}p{1.6cm}>{\centering\arraybackslash}p{1.5cm}>{\centering\arraybackslash}p{1.5cm}|>{\centering\arraybackslash}p{1.6cm}>{\centering\arraybackslash}p{1.5cm}>{\centering\arraybackslash}p{1.5cm}} 
    \toprule
     & & \multicolumn{3}{c|}{\textbf{CIFAR10}} & \multicolumn{3}{c|}{\textbf{CIFAR100}} & \multicolumn{3}{c|}{\textbf{ImageNet200}} & \multicolumn{3}{c}{\textbf{ImageNet1K}} \\
    
     & & Near-OOD & Far-OOD & ID Acc. & Near-OOD & Far-OOD & ID Acc. & Near-OOD & Far-OOD & ID Acc. & Near-OOD & Far-OOD & ID Acc. \\
    \midrule

\multicolumn{7}{l}{\textbf{- Multi-layer Radial Basis Function Networks}} \vspace{.1cm} \\

\multirow{2}{*}{MLRBFN} & DINOv2 & \textcolor{red}{96.01}\textsubscript{\textcolor{red}{(\textpm 0.34)}} & \textcolor{red}{97.60}\textsubscript{\textcolor{red}{(\textpm 0.64)}} & \textcolor{red}{97.79}\textsubscript{\textcolor{red}{(\textpm 0.04)}} & \textcolor{red}{89.18}\textsubscript{\textcolor{red}{(\textpm 0.73)}} & \textcolor{red}{93.18}\textsubscript{\textcolor{red}{(\textpm 1.17)}} & \textcolor{red}{85.25}\textsubscript{\textcolor{red}{(\textpm 0.32)}} & \textcolor{red}{90.24}\textsubscript{\textcolor{red}{(\textpm 0.22)}} & \textcolor{red}{96.82}\textsubscript{\textcolor{red}{(\textpm 0.52)}} & \textcolor{red}{94.61}\textsubscript{\textcolor{red}{(\textpm 0.37)}} & \textcolor{red}{78.84}\textsubscript{\textcolor{red}{(\textpm 0.25)}} & \textcolor{red}{92.57}\textsubscript{\textcolor{red}{(\textpm 0.21)}} & \textcolor{red}{78.62}\textsubscript{\textcolor{red}{(\textpm 0.12)}} \\

& CLIP & \textcolor{red}{95.75}\textsubscript{\textcolor{red}{(\textpm 0.35)}} & \textcolor{red}{98.46}\textsubscript{\textcolor{red}{(\textpm 0.20)}} & \textcolor{red}{95.78}\textsubscript{\textcolor{red}{(\textpm 0.16)}} & \textcolor{red}{86.53}\textsubscript{\textcolor{red}{(\textpm 0.84)}} & \textcolor{red}{95.96}\textsubscript{\textcolor{red}{(\textpm 1.35)}} & \textcolor{red}{76.04}\textsubscript{\textcolor{red}{(\textpm 0.17)}} & \textcolor{red}{85.05}\textsubscript{\textcolor{red}{(\textpm 0.42)}} & \textcolor{red}{94.78}\textsubscript{\textcolor{red}{(\textpm 0.42)}} & \textcolor{red}{90.97}\textsubscript{\textcolor{red}{(\textpm 0.07)}} & \textcolor{red}{72.03}\textsubscript{\textcolor{red}{(\textpm 0.51)}} & \textcolor{red}{88.50}\textsubscript{\textcolor{red}{(\textpm 0.32)}} & \textcolor{red}{72.16}\textsubscript{\textcolor{red}{(\textpm 0.11)}} \\

\multicolumn{7}{l}{\textbf{- Foundation Model Benchmarks}} \vspace{.1cm} \\

\multirow{2}{*}{MSP} & DINOv2 & \textcolor{red}{86.74}\textsubscript{\textcolor{red}{(\textpm 0.50)}} & \textcolor{red}{90.30}\textsubscript{\textcolor{red}{(\textpm 1.59)}} & \textcolor{red}{98.13}\textsubscript{\textcolor{red}{(\textpm 0.09)}} & \textcolor{red}{83.85}\textsubscript{\textcolor{red}{(\textpm 0.46)}} & \textcolor{red}{87.28}\textsubscript{\textcolor{red}{(\textpm 0.88)}} & \textcolor{red}{87.62}\textsubscript{\textcolor{red}{(\textpm 0.17)}} & \textcolor{red}{79.09}\textsubscript{\textcolor{red}{(\textpm 0.33)}} & \textcolor{red}{86.25}\textsubscript{\textcolor{red}{(\textpm 0.18)}} & \textcolor{red}{94.95}\textsubscript{\textcolor{red}{(\textpm 0.08)}} & \textcolor{red}{67.80}\textsubscript{\textcolor{red}{(\textpm 0.38)}} & \textcolor{red}{72.99}\textsubscript{\textcolor{red}{(\textpm 0.46)}} & \textcolor{red}{77.94}\textsubscript{\textcolor{red}{(\textpm 0.17)}} \\

& CLIP & \textcolor{red}{77.67}\textsubscript{\textcolor{red}{(\textpm 0.24)}} & \textcolor{red}{76.56}\textsubscript{\textcolor{red}{(\textpm 1.80)}} & \textcolor{red}{95.92}\textsubscript{\textcolor{red}{(\textpm 0.09)}} & \textcolor{red}{72.85}\textsubscript{\textcolor{red}{(\textpm 0.67)}} & \textcolor{red}{72.08}\textsubscript{\textcolor{red}{(\textpm 1.12)}} & \textcolor{red}{79.14}\textsubscript{\textcolor{red}{(\textpm 0.17)}} & \textcolor{red}{75.55}\textsubscript{\textcolor{red}{(\textpm 0.60)}} & \textcolor{red}{83.51}\textsubscript{\textcolor{red}{(\textpm 0.91)}} & \textcolor{red}{92.73}\textsubscript{\textcolor{red}{(\textpm 0.10)}} & \textcolor{red}{65.04}\textsubscript{\textcolor{red}{(\textpm 0.32)}} & \textcolor{red}{69.84}\textsubscript{\textcolor{red}{(\textpm 0.67)}} & \textcolor{red}{72.21}\textsubscript{\textcolor{red}{(\textpm 0.11)}} \\ \grayline

\multirow{2}{*}{ODIN} & DINOv2 & 96.27\textsubscript{(\textpm 0.27)} & 97.16\textsubscript{(\textpm 0.51)} & 98.13\textsubscript{(\textpm 0.09)} & 85.70\textsubscript{(\textpm 0.54)} & 89.06\textsubscript{(\textpm 1.18)} & 87.62\textsubscript{(\textpm 0.17)} & 81.54\textsubscript{(\textpm 0.48)} & 84.49\textsubscript{(\textpm 0.27)} & 94.96\textsubscript{(\textpm 0.08)} & 55.61\textsubscript{(\textpm 1.70)} & 49.15\textsubscript{(\textpm 2.14)} & 78.42\textsubscript{(\textpm 0.12)} \\

& CLIP & 83.56\textsubscript{(\textpm 1.07)} & 74.32\textsubscript{(\textpm 2.48)} & 95.92\textsubscript{(\textpm 0.09)} & 64.73\textsubscript{(\textpm 1.81)} & 53.04\textsubscript{(\textpm 0.79)} & 79.16\textsubscript{(\textpm 0.17)} & 69.18\textsubscript{(\textpm 1.20)} & 73.42\textsubscript{(\textpm 1.88)} & 92.74\textsubscript{(\textpm 0.10)} & 61.51\textsubscript{(\textpm 0.94)} & 63.16\textsubscript{(\textpm 2.32)} & 72.74\textsubscript{(\textpm 0.15)} \\ \grayline

\multirow{2}{*}{GEN} & DINOv2 & 96.98\textsubscript{(\textpm 0.18)} & 97.85\textsubscript{(\textpm 0.30)} & 98.13\textsubscript{(\textpm 0.09)} & 88.43\textsubscript{(\textpm 0.20)} & 91.73\textsubscript{(\textpm 0.78)} & 87.62\textsubscript{(\textpm 0.17)} & 88.03\textsubscript{(\textpm 0.39)} & 94.06\textsubscript{(\textpm 0.08)} & 94.95\textsubscript{(\textpm 0.08)} & 70.17\textsubscript{(\textpm 0.56)} & 75.17\textsubscript{(\textpm 0.53)} & 77.94\textsubscript{(\textpm 0.17)} \\ 

& CLIP & 89.19\textsubscript{(\textpm 0.23)} & 87.01\textsubscript{(\textpm 1.26)} & 95.92\textsubscript{(\textpm 0.09)} & 74.49\textsubscript{(\textpm 0.65)} & 71.62\textsubscript{(\textpm 0.90)} & 79.14\textsubscript{(\textpm 0.17)} & 80.68\textsubscript{(\textpm 0.68)} & 88.12\textsubscript{(\textpm 0.67)} & 92.73\textsubscript{(\textpm 0.10)} & 68.50\textsubscript{(\textpm 0.40)} & 75.33\textsubscript{(\textpm 1.16)} & 72.21\textsubscript{(\textpm 0.11)} \\ \grayline

\multirow{2}{*}{SHEIP} & DINOv2 & 95.24\textsubscript{(\textpm 0.06)} & 97.49\textsubscript{(\textpm 0.06)} & 98.13\textsubscript{(\textpm 0.09)} & 94.06\textsubscript{(\textpm 0.04)} & 95.49\textsubscript{(\textpm 0.39)} & 87.62\textsubscript{(\textpm 0.17)} & 91.78\textsubscript{(\textpm 0.04)} & 98.30\textsubscript{(\textpm 0.07)} & 94.95\textsubscript{(\textpm 0.08)} & 81.78\textsubscript{(\textpm 0.17)} & 94.83\textsubscript{(\textpm 0.11)} & 77.94\textsubscript{(\textpm 0.17)} \\ 

& CLIP & 93.53\textsubscript{(\textpm 0.07)} & 98.95\textsubscript{(\textpm 0.06)} & 95.92\textsubscript{(\textpm 0.09)} & 88.74\textsubscript{(\textpm 0.18)} & 97.44\textsubscript{(\textpm 0.20)} & 79.14\textsubscript{(\textpm 0.17)} & 80.15\textsubscript{(\textpm 0.17)} & 89.78\textsubscript{(\textpm 0.48)} & 92.73\textsubscript{(\textpm 0.10)} & 66.77\textsubscript{(\textpm 0.44)} & 79.56\textsubscript{(\textpm 0.73)} & 72.21\textsubscript{(\textpm 0.11)} \\ \grayline

\multirow{2}{*}{SHEE} & DINOv2 & 97.62\textsubscript{(\textpm 0.01)} & 98.74\textsubscript{(\textpm 0.05)} & 98.13\textsubscript{(\textpm 0.09)} & 94.89\textsubscript{(\textpm 0.06)} & 96.41\textsubscript{(\textpm 0.07)} & 87.62\textsubscript{(\textpm 0.17)} & 92.60\textsubscript{(\textpm 0.04)} & 97.83\textsubscript{(\textpm 0.08)} & 94.95\textsubscript{(\textpm 0.08)} & 83.58\textsubscript{(\textpm 0.11)} & 94.10\textsubscript{(\textpm 0.07)} & 77.94\textsubscript{(\textpm 0.17)} \\ 

& CLIP & 94.72\textsubscript{(\textpm 0.05)} & 99.10\textsubscript{(\textpm 0.01)} & 95.92\textsubscript{(\textpm 0.09)} & 89.74\textsubscript{(\textpm 0.15)} & 98.03\textsubscript{(\textpm 0.18)} & 79.14\textsubscript{(\textpm 0.17)} & 85.57\textsubscript{(\textpm 0.07)} & 93.41\textsubscript{(\textpm 0.31)} & 92.73\textsubscript{(\textpm 0.10)} & 73.97\textsubscript{(\textpm 0.31)} & 85.44\textsubscript{(\textpm 0.26)} & 72.21\textsubscript{(\textpm 0.11)} \\ \grayline

\multirow{2}{*}{KNN} & DINOv2 & 95.57\textsubscript{(\textpm 0.00)} & 96.06\textsubscript{(\textpm 0.00)} & 98.13\textsubscript{(\textpm 0.09)} & 91.44\textsubscript{(\textpm 0.00)} & 86.57\textsubscript{(\textpm 0.00)} & 87.62\textsubscript{(\textpm 0.17)} & 81.51\textsubscript{(\textpm 0.00)} & 96.20\textsubscript{(\textpm 0.00)} & 94.95\textsubscript{(\textpm 0.08)} & 77.53\textsubscript{(\textpm 0.00)} & 94.65\textsubscript{(\textpm 0.00)} & 77.94\textsubscript{(\textpm 0.17)} \\ 

& CLIP & 90.71\textsubscript{(\textpm 0.00)} & 97.82\textsubscript{(\textpm 0.00)} & 95.92\textsubscript{(\textpm 0.09)} & 82.14\textsubscript{(\textpm 0.00)} & 91.39\textsubscript{(\textpm 0.00)} & 79.14\textsubscript{(\textpm 0.17)} & 68.36\textsubscript{(\textpm 0.00)} & 72.28\textsubscript{(\textpm 0.00)} & 92.73\textsubscript{(\textpm 0.10)} & 57.80\textsubscript{(\textpm 0.00)} & 61.12\textsubscript{(\textpm 0.00)} & 72.21\textsubscript{(\textpm 0.11)} \\ \grayline

\multirow{2}{*}{RMDS} & DINOv2 & 98.27\textsubscript{(\textpm 0.00)} & 99.19\textsubscript{(\textpm 0.00)} & 98.13\textsubscript{(\textpm 0.09)} & 95.16\textsubscript{(\textpm 0.00)} & 97.53\textsubscript{(\textpm 0.00)} & 87.62\textsubscript{(\textpm 0.17)} & 94.77\textsubscript{(\textpm 0.00)} & 98.89\textsubscript{(\textpm 0.00)} & 94.95\textsubscript{(\textpm 0.08)} & 83.79\textsubscript{(\textpm 0.00)} & 96.31\textsubscript{(\textpm 0.00)} & 77.94\textsubscript{(\textpm 0.17)} \\ 

& CLIP & 95.73\textsubscript{(\textpm 0.00)} & 97.81\textsubscript{(\textpm 0.00)} & 95.92\textsubscript{(\textpm 0.09)} & 93.23\textsubscript{(\textpm 0.00)} & 98.81\textsubscript{(\textpm 0.00)} & 79.14\textsubscript{(\textpm 0.17)} & 92.09\textsubscript{(\textpm 0.00)} & 97.88\textsubscript{(\textpm 0.00)} & 92.73\textsubscript{(\textpm 0.10)} & 78.21\textsubscript{(\textpm 0.00)} & 91.50\textsubscript{(\textpm 0.00)} & 72.21\textsubscript{(\textpm 0.11)} \\ 

\bottomrule
    \end{tabular}}\par}
\end{table*}

The results in \autoref{tab:openood_results} are promising; however, since OpenOOD v1.5 uses ResNet18 and ResNet50 architectures without freezing any feature extractor parameters, it is challenging to make a direct comparison between the MLRBFN results and OpenOOD v1.5 benchmarks. To address this issue, we tested multiple methods implemented in OpenOOD v1.5 on features extracted from CLIP and DINOv2. Specifically, we implemented the MSP, ODIN, GEN, SHEIP (SHE-Inner-Product), SHEE (SHE-Euclidean), KNN, and RMDS methods \cite{hendrycks2016baseline,liang2017enhancing,liu2023gen,zhang2022out,ren2021simple,sun2022out}. All implementations were taken from OpenOOD's open-source library. 

We chose to evaluate these methods due to their popularity or relevance to the MLRBFN approach. MSP is the most intuitive method for OOD detection - it simply classifies examples as ID or OOD based on the maximum output probability. With the exception of our modified architecture, this is the approach we take to OOD detection with MLRBFNs, \textit{making the MSP OOD detection baseline most relevant for comparison to MLRBFNs.} ODIN involves temperature scaling and input preprocessing prior to using MSP, and is also quite popular. GEN is similar to ODIN and employs a generalized entropy to transform softmax outputs. Finally, SHEIP, SHEE, RMDS, and KNN each construct prototype representations of ID inputs in the feature space or penultimate layer. These approaches are similar to the philosophy behind MLRBFNs as they attempt to directly model the training data manifold and detect samples which do not lie on it. Finally, we note that since these methods are architecture-agnostic, \emph{with little to no work, each of these methods can be implemented in conjunction with MLRBFNs, further improving OOD detection performance.}

Results are shown in \autoref{tab:foundation_results}. We can see that MLRBFNs outperform MSP on every OOD detection task, implying that \emph{RBF layers are inherently better able to detect OOD inputs than linear layers} given the same feature extractor and feature space. This is the primary goal of the MLRBFN method - to develop a layer structure better able to detect OOD inputs without post-hoc techniques, advanced training methods, or outlier exposure. 

Comparing MLRBFN results with GEN, ODIN, SHEIP, SHEE, KNN, and RMDS demonstrates that all of these methods have comparable OOD detection performance. In most cases, GEN and ODIN outperform MSP but underperform MLRBFNs. Nevertheless, this may indicate that GEN and ODIN's input preprocessing, temperature scaling, and generalized entropy techniques could be used in conjunction with MLRBFNs to further improve OOD detection performance. In many instances, SHEIP, SHEE, KNN, or RMDS have similar or better performance on OOD detection compared to MLRBFNs. This means that the features extracted from foundation models can be used to classify inputs as ID or OOD, and suggests that foundation models are less prone to feature collapse. It also explains why these methods did not perform extremely well in the OpenOOD v1.5 study, which trained a feature extractor on the ID classification problem, but work much better when using feature extractors that are also effective on images outside the ID data. Again, we note that these techniques are not mutually exclusive with MLRBFNs, and it is likely that an OOD detector could achieve superior performance by using an ensemble of an MLRBFN with post-hoc processors with features extracted by a foundation model. 

\subsubsection{OOD Detection with Increasing Number of Classification Layers}

\begin{figure}[t]
    \centering
    \includegraphics[width=\linewidth]{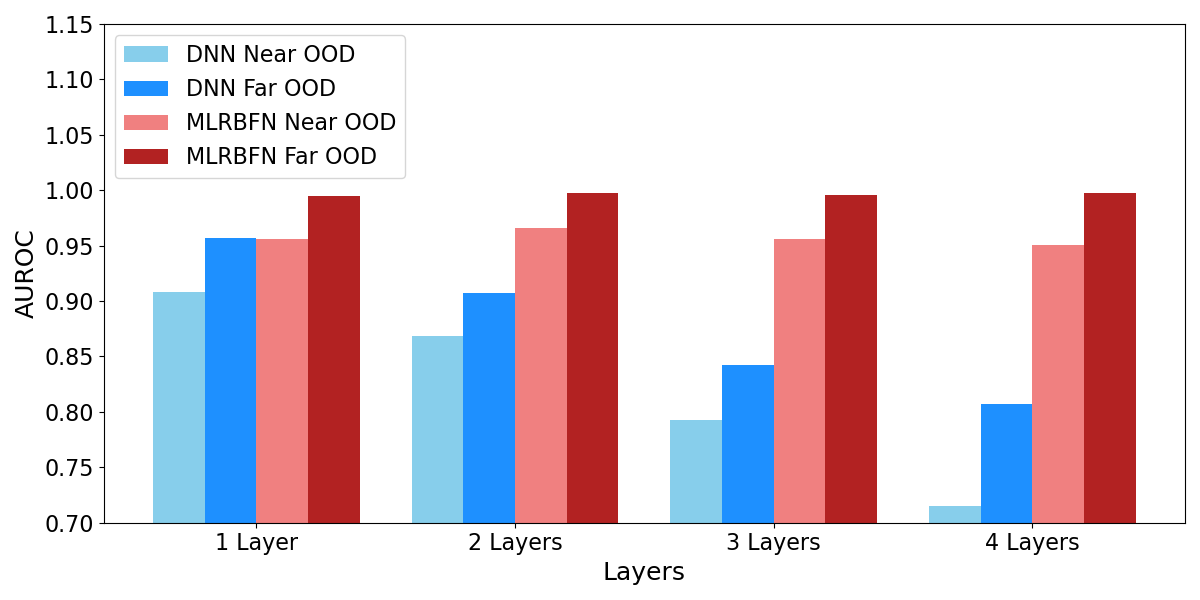} 
    \caption{AUROC for near- and far-OOD tasks on the CIFAR10 dataset. MSP was used for DNN OOD detection. As the number of classification layers increases, MLRBFNs maintain high AUROC for the OOD detection task, while DNNs have decreasing AUROC. This supports MLRBFNs as networks which are naturally capable of detecting OOD inputs.}
    \label{fig:CLIP_number_of_layer_LC_vs_RBF}
\end{figure}

Finally, we tested how the number of trainable ``head'' layers affects OOD detection performance in DNNs and MLRBFNs. To do this, we trained networks with 1, 2, 3, or 4 head layers on CIFAR10 features extracted by CLIP, and measured AUROC on near-OOD and far-OOD datasets. Results are shown in \autoref{fig:CLIP_number_of_layer_LC_vs_RBF}. As can be seen, MLRBFN performance slightly increases as the number of layers increases, while DNNs with ReLU activation functions produce worse performance as the number of layers increases. This may be a sign that feature collapse can occur rapidly, as the DNN's trainable ``head'' layers map all inputs to one of the training classes even when DNNs use powerful frozen feature extractors. MLRBFNs are able to achieve similar accuracy to DNNs but do not have degrading robustness to OOD inputs. 
\section{Conclusion}

We have developed the first method for directly training multi-layer RBF networks using backpropagation for effective OOD detection by identifying key failure cases that required remediation. A primary issue is the $\mathbf{0}$-mapped class, where the zero-vector response would be mapped to one class that occurred ``everywhere'' and inhibited OOD detection. We resolved this challenge by introducing a novel depression mechanism to carry the ``OOD-ness'' of an input forward to the following layers. In doing so we show a new approach to training MLRBFN classification heads that directly tackles classification and OOD detection simultaneously. MLRBFNs are competitive with state-of-the-art methods and perform significantly better than the MSP baseline. This architecture opens a new avenue of research directions on prior methods that may be possible to effectively combine with modern deep learning techniques. We hope this work inspires further interest in building neural network layers which are inherently capable of modeling the training data manifold and identifying OOD inputs.

\label{sec:conclusion}
{
    \small
    \bibliographystyle{ieeenat_fullname}
    \bibliography{main}
}

\clearpage
\setcounter{page}{1}
\maketitlesupplementary

\section{Initialization Details}
\label{sec:initialization}

For layer $\ell$ with $N$ centroids, $k$-means clustering with $N$ centroids was used to find the initialization positions for the centroids. This can be done quickly for a large batch of data using web-scale $k$-means clustering with $k$-means++ initialization \cite{sculley2010web,arthur2006k}. Then for each datapoint in the initialization batch, the distance to its closest centroid is computed, producing a vector in $\mathbb{R}^n$, where $n$ is the number of datapoints in the batch. Similarly, for each centroid, the distance to its closest datapoint was computed, producing a vector in $\mathbb{R}^N$. Note that distances were computed using $\lVert \cdot \rVert_k$, which is also used in the RBFN forward pass. 

To set $\beta_{c(\ell)}^+$, we take inspiration from the Gaussian distribution. In the Gaussian distribution, 95\% of datapoints fall within 2 standard deviations of the mean. This can be represented as $2\sigma = d$, or equivalently, $4 \sigma^2 = d^2$. Since $\frac{1}{\sigma^2} \propto \beta_{c(\ell)}^+$, we should expect to set $\beta_{c(\ell)}^+$ proportional to $\frac{4}{d^2}$. 

Returning to our two vectors in $\mathbb{R}^n$ and $\mathbb{R}^N$, call the larger of the 95\% quantile of each of these metrics $d$. We can use this $d$ as an estimate of the necessary width of a Gaussian distribution, and thus set $\beta_{c(\ell)}^+ = \frac{4}{d^2}$. Note that we initialize every inverse-width in a layer to the same value, $\frac{4}{d^2}$. In our code, where $\beta_{c(\ell)}^+ = \texttt{softplus}(\beta_{c(\ell)})$, we actually initialize $\beta_{c(\ell)} = \texttt{inverse-softplus}(\frac{4}{d^2})$.

To see this procedure work in code, please see our code in \autoref{sec:code}. 
\section{Moons}
\label{sec:moons}

To enable development and achieve a simple baseline, we employed a 4-class moons problem. Similar datasets have been used in other papers on uncertainty quantification due to their ease of visualization. 

The 4-class moons dataset was generated from scikit-learn's make moons function with noise parameter 0.2, which produces a 2-class moons problem \cite{scikit-learn}. Two extra classes were generated by shifting the base 2-class moons problem by 2 units in each direction. Overall, the dataset had 1000 training datapoints and 500 testing datapoints split equally among each class. The dataset was normalized prior to training. 

We trained a 3-layer MLRBFN with $N = 50$ centroids in the first and second layers and $N = C = 4$ centroids in the third layer. A projection dimension of $o = 100$ was used in the first and second layers. $k$ was set to 2 for all layers. The Adam optimizer was used with a learning rate of $10^{-3}$ \cite{kingma2014adam}. The network was trained for 250 epochs with a batch size of 100. 

\autoref{fig:teaser} demonstrates the performance of our MLRBFN on the 4-class moons problem. Datapoints are scattered and colored according to their true class. Color in the background represents the class prediction while shading represents the confidence of the network, with white being low confidence and dark being high confidence. It is clear that the network can classify the dataset correctly ($>99.5$\% training and testing accuracy) and is only confident near the training data manifold. 
\section{Additional Results}
\label{sec:add_results}

In this section, we include additional results for AUPR Out and FPR@95 metrics. These can be seen in \autoref{tab:aupr_results} and \autoref{tab:fpr_results}, respectively. 

From \autoref{tab:aupr_results}, it is clear that MLRBFNs produce significantly better results than MSP and are competitive with other presented methods. Some other methods, like RMDS, have significantly better performance across all datasets, but these methods can be incorporated with MLRBFNs to produce even better performance. 

From \autoref{tab:fpr_results}, MLRBFN results are significantly better than MSP: MSP has a 100\% false positive rate across all datasets, while MLRBFNs reduce this significantly. Again, MLRBFNs can be combined with other methods presented in the table to produce improved performance. 

\begin{table*}[!ht]
\caption{Comparing the performance of MLRBFNs to published benchmarks and existing methods. In this table, we use AUPR Out as the metric for OOD detection. Whenever applicable, we report the average number and the corresponding standard deviation obtained from 3 training runs.}
\label{tab:aupr_results}
\centering
{\renewcommand\baselinestretch{1.3}\selectfont\resizebox{\textwidth}{!}{
    \begin{tabular}{ll|>{\centering\arraybackslash}p{1.6cm}>{\centering\arraybackslash}p{1.5cm}>{\centering\arraybackslash}p{1.5cm}|>{\centering\arraybackslash}p{1.6cm}>{\centering\arraybackslash}p{1.5cm}>{\centering\arraybackslash}p{1.5cm}|>{\centering\arraybackslash}p{1.6cm}>{\centering\arraybackslash}p{1.5cm}>{\centering\arraybackslash}p{1.5cm}|>{\centering\arraybackslash}p{1.6cm}>{\centering\arraybackslash}p{1.5cm}>{\centering\arraybackslash}p{1.5cm}} 
    \toprule
     & & \multicolumn{3}{c|}{\textbf{CIFAR10}} & \multicolumn{3}{c|}{\textbf{CIFAR100}} & \multicolumn{3}{c|}{\textbf{ImageNet200}} & \multicolumn{3}{c}{\textbf{ImageNet1K}} \\
    
     & & Near-OOD & Far-OOD & ID Acc. & Near-OOD & Far-OOD & ID Acc. & Near-OOD & Far-OOD & ID Acc. & Near-OOD & Far-OOD & ID Acc. \\
    \midrule

\multicolumn{7}{l}{\textbf{- Multi-layer Radial Basis Function Networks}} \vspace{.1cm} \\

\multirow{2}{*}{MLRBFN} & DINOv2 & \textcolor{red}{95.77}\textsubscript{\textcolor{red}{(\textpm 0.41)}} & \textcolor{red}{98.20}\textsubscript{\textcolor{red}{(\textpm 0.60)}} & \textcolor{red}{97.79}\textsubscript{\textcolor{red}{(\textpm 0.04)}} & \textcolor{red}{86.47}\textsubscript{\textcolor{red}{(\textpm 1.13)}} & \textcolor{red}{93.80}\textsubscript{\textcolor{red}{(\textpm 1.23)}} & \textcolor{red}{85.25}\textsubscript{\textcolor{red}{(\textpm 0.32)}} & \textcolor{red}{93.50}\textsubscript{\textcolor{red}{(\textpm 0.17)}} & \textcolor{red}{95.83}\textsubscript{\textcolor{red}{(\textpm 0.62)}} & \textcolor{red}{94.61}\textsubscript{\textcolor{red}{(\textpm 0.37)}} & \textcolor{red}{58.82}\textsubscript{\textcolor{red}{(\textpm 1.07)}} & \textcolor{red}{69.61}\textsubscript{\textcolor{red}{(\textpm 0.62)}} & \textcolor{red}{78.62}\textsubscript{\textcolor{red}{(\textpm 0.12)}}
 \\

& CLIP & \textcolor{red}{95.27}\textsubscript{\textcolor{red}{(\textpm 0.51)}} & \textcolor{red}{98.65}\textsubscript{\textcolor{red}{(\textpm 0.30)}} & \textcolor{red}{95.78}\textsubscript{\textcolor{red}{(\textpm 0.16)}} & \textcolor{red}{83.36}\textsubscript{\textcolor{red}{(\textpm 1.41)}} & \textcolor{red}{95.92}\textsubscript{\textcolor{red}{(\textpm 1.70)}} & \textcolor{red}{76.04}\textsubscript{\textcolor{red}{(\textpm 0.17)}} & \textcolor{red}{88.56}\textsubscript{\textcolor{red}{(\textpm 0.24)}} & \textcolor{red}{93.55}\textsubscript{\textcolor{red}{(\textpm 0.76)}} & \textcolor{red}{90.97}\textsubscript{\textcolor{red}{(\textpm 0.07)}} & \textcolor{red}{47.53}\textsubscript{\textcolor{red}{(\textpm 0.91)}} & \textcolor{red}{61.72}\textsubscript{\textcolor{red}{(\textpm 1.21)}} & \textcolor{red}{72.16}\textsubscript{\textcolor{red}{(\textpm 0.11)}}
 \\

\multicolumn{7}{l}{\textbf{- Foundation Model Benchmarks}} \vspace{.1cm} \\

\multirow{2}{*}{MSP} & DINOv2 & \textcolor{red}{91.17}\textsubscript{\textcolor{red}{(\textpm 0.35)}} & \textcolor{red}{94.80}\textsubscript{\textcolor{red}{(\textpm 0.68)}} & \textcolor{red}{98.13}\textsubscript{\textcolor{red}{(\textpm 0.09)}} & \textcolor{red}{78.91}\textsubscript{\textcolor{red}{(\textpm 0.65)}} & \textcolor{red}{87.86}\textsubscript{\textcolor{red}{(\textpm 0.87)}} & \textcolor{red}{87.62}\textsubscript{\textcolor{red}{(\textpm 0.17)}} & \textcolor{red}{88.75}\textsubscript{\textcolor{red}{(\textpm 0.23)}} & \textcolor{red}{88.85}\textsubscript{\textcolor{red}{(\textpm 0.09)}} & \textcolor{red}{94.95}\textsubscript{\textcolor{red}{(\textpm 0.08)}} & \textcolor{red}{43.90}\textsubscript{\textcolor{red}{(\textpm 0.31)}} & \textcolor{red}{34.34}\textsubscript{\textcolor{red}{(\textpm 0.45)}} & \textcolor{red}{77.94}\textsubscript{\textcolor{red}{(\textpm 0.17)}} \\

& CLIP & \textcolor{red}{82.40}\textsubscript{\textcolor{red}{(\textpm 0.15)}} & \textcolor{red}{85.64}\textsubscript{\textcolor{red}{(\textpm 0.97)}} & \textcolor{red}{95.92}\textsubscript{\textcolor{red}{(\textpm 0.09)}} & \textcolor{red}{65.07}\textsubscript{\textcolor{red}{(\textpm 0.48)}} & \textcolor{red}{75.14}\textsubscript{\textcolor{red}{(\textpm 0.76)}} & \textcolor{red}{79.14}\textsubscript{\textcolor{red}{(\textpm 0.17)}} & \textcolor{red}{83.38}\textsubscript{\textcolor{red}{(\textpm 0.29)}} & \textcolor{red}{83.36}\textsubscript{\textcolor{red}{(\textpm 0.51)}} & \textcolor{red}{92.73}\textsubscript{\textcolor{red}{(\textpm 0.10)}} & \textcolor{red}{40.26}\textsubscript{\textcolor{red}{(\textpm 0.24)}} & \textcolor{red}{29.96}\textsubscript{\textcolor{red}{(\textpm 0.54)}} & \textcolor{red}{72.21}\textsubscript{\textcolor{red}{(\textpm 0.11)}} \\ \grayline

\multirow{2}{*}{ODIN} & DINOv2 & 95.24\textsubscript{(\textpm 0.48)} & 96.94\textsubscript{(\textpm 0.68)} & 98.13\textsubscript{(\textpm 0.09)} & 80.47\textsubscript{(\textpm 0.54)} & 88.82\textsubscript{(\textpm 1.41)} & 87.62\textsubscript{(\textpm 0.17)} & 84.27\textsubscript{(\textpm 0.16)} & 79.72\textsubscript{(\textpm 0.46)} & 94.96\textsubscript{(\textpm 0.08)} & 33.94\textsubscript{(\textpm 0.89)} & 16.96\textsubscript{(\textpm 0.69)} & 78.42\textsubscript{(\textpm 0.12)} \\

& CLIP & 80.52\textsubscript{(\textpm 1.62)} & 77.14\textsubscript{(\textpm 2.07)} & 95.92\textsubscript{(\textpm 0.09)} & 57.75\textsubscript{(\textpm 1.48)} & 62.89\textsubscript{(\textpm 0.44)} & 79.16\textsubscript{(\textpm 0.17)} & 73.26\textsubscript{(\textpm 0.85)} & 66.92\textsubscript{(\textpm 1.71)} & 92.74\textsubscript{(\textpm 0.10)} & 36.44\textsubscript{(\textpm 0.97)} & 23.54\textsubscript{(\textpm 1.62)} & 72.74\textsubscript{(\textpm 0.15)} \\ \grayline

\multirow{2}{*}{GEN} & DINOv2 & 96.30\textsubscript{(\textpm 0.24)} & 97.93\textsubscript{(\textpm 0.33)} & 98.13\textsubscript{(\textpm 0.09)} & 84.12\textsubscript{(\textpm 0.32)} & 91.88\textsubscript{(\textpm 0.94)} & 87.62\textsubscript{(\textpm 0.17)} & 91.45\textsubscript{(\textpm 0.14)} & 92.34\textsubscript{(\textpm 0.12)} & 94.95\textsubscript{(\textpm 0.08)} & 43.91\textsubscript{(\textpm 0.81)} & 33.94\textsubscript{(\textpm 0.77)} & 77.94\textsubscript{(\textpm 0.17)} \\ 

& CLIP & 87.69\textsubscript{(\textpm 0.20)} & 88.66\textsubscript{(\textpm 1.00)} & 95.92\textsubscript{(\textpm 0.09)} & 64.77\textsubscript{(\textpm 0.51)} & 73.29\textsubscript{(\textpm 0.48)} & 79.14\textsubscript{(\textpm 0.17)} & 84.16\textsubscript{(\textpm 0.57)} & 85.00\textsubscript{(\textpm 1.00)} & 92.73\textsubscript{(\textpm 0.10)} & 42.15\textsubscript{(\textpm 0.45)} & 35.34\textsubscript{(\textpm 1.83)} & 72.21\textsubscript{(\textpm 0.11)} \\ \grayline

\multirow{2}{*}{SHEIP} & DINOv2 & 95.01\textsubscript{(\textpm 0.08)} & 98.20\textsubscript{(\textpm 0.09)} & 98.13\textsubscript{(\textpm 0.09)} & 93.60\textsubscript{(\textpm 0.08)} & 97.05\textsubscript{(\textpm 0.28)} & 87.62\textsubscript{(\textpm 0.17)} & 94.45\textsubscript{(\textpm 0.08)} & 97.71\textsubscript{(\textpm 0.18)} & 94.95\textsubscript{(\textpm 0.08)} & 63.19\textsubscript{(\textpm 0.55)} & 76.19\textsubscript{(\textpm 0.40)} & 77.94\textsubscript{(\textpm 0.17)} \\ 

& CLIP & 93.08\textsubscript{(\textpm 0.07)} & 98.69\textsubscript{(\textpm 0.05)} & 95.92\textsubscript{(\textpm 0.09)} & 87.61\textsubscript{(\textpm 0.23)} & 97.50\textsubscript{(\textpm 0.24)} & 79.14\textsubscript{(\textpm 0.17)} & 81.93\textsubscript{(\textpm 0.28)} & 85.46\textsubscript{(\textpm 0.62)} & 92.73\textsubscript{(\textpm 0.10)} & 41.70\textsubscript{(\textpm 0.37)} & 37.70\textsubscript{(\textpm 1.12)} & 72.21\textsubscript{(\textpm 0.11)} \\ \grayline

\multirow{2}{*}{SHEE} & DINOv2 & 97.65\textsubscript{(\textpm 0.02)} & 99.30\textsubscript{(\textpm 0.05)} & 98.13\textsubscript{(\textpm 0.09)} & 94.06\textsubscript{(\textpm 0.09)} & 97.30\textsubscript{(\textpm 0.06)} & 87.62\textsubscript{(\textpm 0.17)} & 94.44\textsubscript{(\textpm 0.05)} & 96.62\textsubscript{(\textpm 0.20)} & 94.95\textsubscript{(\textpm 0.08)} & 64.35\textsubscript{(\textpm 0.53)} & 70.89\textsubscript{(\textpm 0.62)} & 77.94\textsubscript{(\textpm 0.17)} \\ 

& CLIP & 94.11\textsubscript{(\textpm 0.06)} & 98.76\textsubscript{(\textpm 0.03)} & 95.92\textsubscript{(\textpm 0.09)} & 88.66\textsubscript{(\textpm 0.24)} & 97.64\textsubscript{(\textpm 0.28)} & 79.14\textsubscript{(\textpm 0.17)} & 86.44\textsubscript{(\textpm 0.26)} & 90.38\textsubscript{(\textpm 0.46)} & 92.73\textsubscript{(\textpm 0.10)} & 47.25\textsubscript{(\textpm 0.39)} & 46.42\textsubscript{(\textpm 0.73)} & 72.21\textsubscript{(\textpm 0.11)} \\ \grayline

\multirow{2}{*}{KNN} & DINOv2 & 95.37\textsubscript{(\textpm 0.00)} & 97.76\textsubscript{(\textpm 0.00)} & 98.13\textsubscript{(\textpm 0.09)} & 91.23\textsubscript{(\textpm 0.00)} & 91.47\textsubscript{(\textpm 0.00)} & 87.62\textsubscript{(\textpm 0.17)} & 86.08\textsubscript{(\textpm 0.00)} & 96.14\textsubscript{(\textpm 0.00)} & 94.95\textsubscript{(\textpm 0.08)} & 60.15\textsubscript{(\textpm 0.00)} & 80.16\textsubscript{(\textpm 0.00)} & 77.94\textsubscript{(\textpm 0.17)} \\ 

& CLIP & 90.51\textsubscript{(\textpm 0.00)} & 97.64\textsubscript{(\textpm 0.00)} & 95.92\textsubscript{(\textpm 0.09)} & 82.59\textsubscript{(\textpm 0.00)} & 91.52\textsubscript{(\textpm 0.00)} & 79.14\textsubscript{(\textpm 0.17)} & 69.15\textsubscript{(\textpm 0.00)} & 61.81\textsubscript{(\textpm 0.00)} & 92.73\textsubscript{(\textpm 0.10)} & 35.04\textsubscript{(\textpm 0.00)} & 21.54\textsubscript{(\textpm 0.00)} & 72.21\textsubscript{(\textpm 0.11)} \\ \grayline

\multirow{2}{*}{RMDS} & DINOv2 & 97.89\textsubscript{(\textpm 0.00)} & 99.17\textsubscript{(\textpm 0.00)} & 98.13\textsubscript{(\textpm 0.09)} & 94.14\textsubscript{(\textpm 0.00)} & 98.29\textsubscript{(\textpm 0.00)} & 87.62\textsubscript{(\textpm 0.17)} & 97.16\textsubscript{(\textpm 0.00)} & 98.63\textsubscript{(\textpm 0.00)} & 94.95\textsubscript{(\textpm 0.08)} & 74.46\textsubscript{(\textpm 0.00)} & 85.21\textsubscript{(\textpm 0.00)} & 77.94\textsubscript{(\textpm 0.17)} \\ 

& CLIP & 94.79\textsubscript{(\textpm 0.00)} & 97.08\textsubscript{(\textpm 0.00)} & 95.92\textsubscript{(\textpm 0.09)} & 92.57\textsubscript{(\textpm 0.00)} & 98.91\textsubscript{(\textpm 0.00)} & 79.14\textsubscript{(\textpm 0.17)} & 94.45\textsubscript{(\textpm 0.00)} & 97.27\textsubscript{(\textpm 0.00)} & 92.73\textsubscript{(\textpm 0.10)} & 55.24\textsubscript{(\textpm 0.00)} & 63.57\textsubscript{(\textpm 0.00)} & 72.21\textsubscript{(\textpm 0.11)} \\ 

\bottomrule
    \end{tabular}}\par}
\end{table*}

\begin{table*}[!ht]
\caption{Comparing the performance of MLRBFNs to published benchmarks and existing methods. In this table, we use FPR@95 (False Positive Rate at 95\% True Positive Rate) as the metric for OOD detection. As such, unlike the previous tables, smaller numbers indicate better performance here. Whenever applicable, we report the average number and the corresponding standard deviation obtained from 3 training runs.}
\label{tab:fpr_results}
\centering
{\renewcommand\baselinestretch{1.3}\selectfont\resizebox{\textwidth}{!}{
    \begin{tabular}{ll|>{\centering\arraybackslash}p{1.6cm}>{\centering\arraybackslash}p{1.5cm}>{\centering\arraybackslash}p{1.5cm}|>{\centering\arraybackslash}p{1.6cm}>{\centering\arraybackslash}p{1.5cm}>{\centering\arraybackslash}p{1.5cm}|>{\centering\arraybackslash}p{1.6cm}>{\centering\arraybackslash}p{1.5cm}>{\centering\arraybackslash}p{1.5cm}|>{\centering\arraybackslash}p{1.6cm}>{\centering\arraybackslash}p{1.5cm}>{\centering\arraybackslash}p{1.5cm}} 
    \toprule
     & & \multicolumn{3}{c|}{\textbf{CIFAR10}} & \multicolumn{3}{c|}{\textbf{CIFAR100}} & \multicolumn{3}{c|}{\textbf{ImageNet200}} & \multicolumn{3}{c}{\textbf{ImageNet1K}} \\
    
     & & Near-OOD & Far-OOD & ID Acc. & Near-OOD & Far-OOD & ID Acc. & Near-OOD & Far-OOD & ID Acc. & Near-OOD & Far-OOD & ID Acc. \\
    \midrule

\multicolumn{7}{l}{\textbf{- Multi-layer Radial Basis Function Networks}} \vspace{.1cm} \\

\multirow{2}{*}{MLRBFN} & DINOv2 & \textcolor{red}{17.99}\textsubscript{\textcolor{red}{(\textpm 2.08)}} & \textcolor{red}{9.51}\textsubscript{\textcolor{red}{(\textpm 2.08)}} & \textcolor{red}{97.79}\textsubscript{\textcolor{red}{(\textpm 0.04)}} & \textcolor{red}{43.39}\textsubscript{\textcolor{red}{(\textpm 2.41)}} & \textcolor{red}{25.31}\textsubscript{\textcolor{red}{(\textpm 4.31)}} & \textcolor{red}{85.25}\textsubscript{\textcolor{red}{(\textpm 0.32)}} & \textcolor{red}{49.55}\textsubscript{\textcolor{red}{(\textpm 1.07)}} & \textcolor{red}{11.71}\textsubscript{\textcolor{red}{(\textpm 2.13)}} & \textcolor{red}{94.61}\textsubscript{\textcolor{red}{(\textpm 0.37)}} & \textcolor{red}{67.08}\textsubscript{\textcolor{red}{(\textpm 1.03)}} & \textcolor{red}{26.92}\textsubscript{\textcolor{red}{(\textpm 1.48)}} & \textcolor{red}{78.62}\textsubscript{\textcolor{red}{(\textpm 0.12)}} \\

& CLIP & \textcolor{red}{16.81}\textsubscript{\textcolor{red}{(\textpm 1.78)}} & \textcolor{red}{6.36}\textsubscript{\textcolor{red}{(\textpm 1.30)}} & \textcolor{red}{95.78}\textsubscript{\textcolor{red}{(\textpm 0.16)}} & \textcolor{red}{46.16}\textsubscript{\textcolor{red}{(\textpm 1.87)}} & \textcolor{red}{14.21}\textsubscript{\textcolor{red}{(\textpm 3.57)}} & \textcolor{red}{76.04}\textsubscript{\textcolor{red}{(\textpm 0.17)}} & \textcolor{red}{60.18}\textsubscript{\textcolor{red}{(\textpm 1.89)}} & \textcolor{red}{22.03}\textsubscript{\textcolor{red}{(\textpm 2.23)}} & \textcolor{red}{90.97}\textsubscript{\textcolor{red}{(\textpm 0.07)}} & \textcolor{red}{76.68}\textsubscript{\textcolor{red}{(\textpm 1.06)}} & \textcolor{red}{45.61}\textsubscript{\textcolor{red}{(\textpm 2.35)}} & \textcolor{red}{72.16}\textsubscript{\textcolor{red}{(\textpm 0.11)}} \\

\multicolumn{7}{l}{\textbf{- Foundation Model Benchmarks}} \vspace{.1cm} \\

\multirow{2}{*}{MSP} & DINOv2 & \textcolor{red}{100.00}\textsubscript{\textcolor{red}{(\textpm 0.00)}} & \textcolor{red}{100.00}\textsubscript{\textcolor{red}{(\textpm 0.00)}} & \textcolor{red}{98.13}\textsubscript{\textcolor{red}{(\textpm 0.09)}} & \textcolor{red}{100.00}\textsubscript{\textcolor{red}{(\textpm 0.00)}} & \textcolor{red}{45.78}\textsubscript{\textcolor{red}{(\textpm 1.41)}} & \textcolor{red}{87.62}\textsubscript{\textcolor{red}{(\textpm 0.17)}} & \textcolor{red}{100.00}\textsubscript{\textcolor{red}{(\textpm 0.00)}} & \textcolor{red}{100.00}\textsubscript{\textcolor{red}{(\textpm 0.00)}} & \textcolor{red}{94.95}\textsubscript{\textcolor{red}{(\textpm 0.08)}} & \textcolor{red}{100.00}\textsubscript{\textcolor{red}{(\textpm 0.00)}} & \textcolor{red}{100.00}\textsubscript{\textcolor{red}{(\textpm 0.00)}} & \textcolor{red}{77.94}\textsubscript{\textcolor{red}{(\textpm 0.17)}} \\

& CLIP & \textcolor{red}{100.00}\textsubscript{\textcolor{red}{(\textpm 0.00)}} & \textcolor{red}{100.00}\textsubscript{\textcolor{red}{(\textpm 0.00)}} & \textcolor{red}{95.92}\textsubscript{\textcolor{red}{(\textpm 0.09)}} & \textcolor{red}{100.00}\textsubscript{\textcolor{red}{(\textpm 0.00)}} & \textcolor{red}{77.12}\textsubscript{\textcolor{red}{(\textpm 7.57)}} & \textcolor{red}{79.14}\textsubscript{\textcolor{red}{(\textpm 0.17)}} & \textcolor{red}{100.00}\textsubscript{\textcolor{red}{(\textpm 0.00)}} & \textcolor{red}{100.00}\textsubscript{\textcolor{red}{(\textpm 0.00)}} & \textcolor{red}{92.73}\textsubscript{\textcolor{red}{(\textpm 0.10)}} & \textcolor{red}{100.00}\textsubscript{\textcolor{red}{(\textpm 0.00)}} & \textcolor{red}{100.00}\textsubscript{\textcolor{red}{(\textpm 0.00)}} & \textcolor{red}{72.21}\textsubscript{\textcolor{red}{(\textpm 0.11)}} \\ \grayline

\multirow{2}{*}{ODIN} & DINOv2 & 14.11\textsubscript{(\textpm 0.87)} & 10.78\textsubscript{(\textpm 1.71)} & 98.13\textsubscript{(\textpm 0.09)} & 49.59\textsubscript{(\textpm 2.62)} & 33.50\textsubscript{(\textpm 2.68)} & 87.62\textsubscript{(\textpm 0.17)} & 62.22\textsubscript{(\textpm 2.24)} & 47.14\textsubscript{(\textpm 1.25)} & 94.96\textsubscript{(\textpm 0.08)} & 87.52\textsubscript{(\textpm 0.61)} & 88.99\textsubscript{(\textpm 1.36)} & 78.42\textsubscript{(\textpm 0.12)} \\

& CLIP & 55.69\textsubscript{(\textpm 2.25)} & 65.67\textsubscript{(\textpm 5.08)} & 95.92\textsubscript{(\textpm 0.09)} & 86.48\textsubscript{(\textpm 3.18)} & 75.28\textsubscript{(\textpm 1.54)} & 79.16\textsubscript{(\textpm 0.17)} & 77.12\textsubscript{(\textpm 1.43)} & 67.04\textsubscript{(\textpm 2.91)} & 92.74\textsubscript{(\textpm 0.10)} & 82.37\textsubscript{(\textpm 1.23)} & 78.30\textsubscript{(\textpm 1.81)} & 72.74\textsubscript{(\textpm 0.15)} \\ \grayline

\multirow{2}{*}{GEN} & DINOv2 & 11.56\textsubscript{(\textpm 0.89)} & 8.32\textsubscript{(\textpm 1.16)} & 98.13\textsubscript{(\textpm 0.09)} & 42.56\textsubscript{(\textpm 0.82)} & 27.86\textsubscript{(\textpm 2.36)} & 87.62\textsubscript{(\textpm 0.17)} & 54.84\textsubscript{(\textpm 3.23)} & 24.37\textsubscript{(\textpm 0.28)} & 94.95\textsubscript{(\textpm 0.08)} & 76.21\textsubscript{(\textpm 1.17)} & 67.28\textsubscript{(\textpm 0.99)} & 77.94\textsubscript{(\textpm 0.17)} \\ 

& CLIP & 46.15\textsubscript{(\textpm 1.00)} & 53.18\textsubscript{(\textpm 5.71)} & 95.92\textsubscript{(\textpm 0.09)} & 67.44\textsubscript{(\textpm 2.49)} & 65.24\textsubscript{(\textpm 2.82)} & 79.14\textsubscript{(\textpm 0.17)} & 66.89\textsubscript{(\textpm 1.49)} & 44.34\textsubscript{(\textpm 2.10)} & 92.73\textsubscript{(\textpm 0.10)} & 76.47\textsubscript{(\textpm 0.81)} & 67.22\textsubscript{(\textpm 1.64)} & 72.21\textsubscript{(\textpm 0.11)} \\ \grayline

\multirow{2}{*}{SHEIP} & DINOv2 & 18.00\textsubscript{(\textpm 0.08)} & 7.85\textsubscript{(\textpm 0.18)} & 98.13\textsubscript{(\textpm 0.09)} & 24.63\textsubscript{(\textpm 0.25)} & 14.80\textsubscript{(\textpm 0.71)} & 87.62\textsubscript{(\textpm 0.17)} & 43.91\textsubscript{(\textpm 0.28)} & 5.70\textsubscript{(\textpm 0.18)} & 94.95\textsubscript{(\textpm 0.08)} & 65.63\textsubscript{(\textpm 0.46)} & 20.64\textsubscript{(\textpm 0.88)} & 77.94\textsubscript{(\textpm 0.17)} \\ 

& CLIP & 23.36\textsubscript{(\textpm 0.15)} & 4.52\textsubscript{(\textpm 0.20)} & 95.92\textsubscript{(\textpm 0.09)} & 32.41\textsubscript{(\textpm 0.51)} & 10.30\textsubscript{(\textpm 0.83)} & 79.14\textsubscript{(\textpm 0.17)} & 60.18\textsubscript{(\textpm 0.39)} & 32.65\textsubscript{(\textpm 0.81)} & 92.73\textsubscript{(\textpm 0.10)} & 80.41\textsubscript{(\textpm 0.52)} & 55.66\textsubscript{(\textpm 1.29)} & 72.21\textsubscript{(\textpm 0.11)} \\ \grayline

\multirow{2}{*}{SHEE} & DINOv2 & 10.09\textsubscript{(\textpm 0.06)} & 4.22\textsubscript{(\textpm 0.13)} & 98.13\textsubscript{(\textpm 0.09)} & 21.16\textsubscript{(\textpm 0.04)} & 11.03\textsubscript{(\textpm 0.27)} & 87.62\textsubscript{(\textpm 0.17)} & 40.78\textsubscript{(\textpm 0.13)} & 6.19\textsubscript{(\textpm 0.15)} & 94.95\textsubscript{(\textpm 0.08)} & 59.54\textsubscript{(\textpm 0.35)} & 20.29\textsubscript{(\textpm 0.38)} & 77.94\textsubscript{(\textpm 0.17)} \\ 

& CLIP & 18.93\textsubscript{(\textpm 0.13)} & 3.62\textsubscript{(\textpm 0.03)} & 95.92\textsubscript{(\textpm 0.09)} & 31.17\textsubscript{(\textpm 0.37)} & 6.79\textsubscript{(\textpm 0.52)} & 79.14\textsubscript{(\textpm 0.17)} & 50.88\textsubscript{(\textpm 0.13)} & 25.04\textsubscript{(\textpm 0.58)} & 92.73\textsubscript{(\textpm 0.10)} & 69.66\textsubscript{(\textpm 0.21)} & 42.95\textsubscript{(\textpm 0.49)} & 72.21\textsubscript{(\textpm 0.11)} \\ \grayline

\multirow{2}{*}{KNN} & DINOv2 & 15.25\textsubscript{(\textpm 0.00)} & 10.92\textsubscript{(\textpm 0.00)} & 98.13\textsubscript{(\textpm 0.09)} & 33.12\textsubscript{(\textpm 0.00)} & 27.77\textsubscript{(\textpm 0.00)} & 87.62\textsubscript{(\textpm 0.17)} & 66.97\textsubscript{(\textpm 0.00)} & 19.31\textsubscript{(\textpm 0.00)} & 94.95\textsubscript{(\textpm 0.08)} & 71.93\textsubscript{(\textpm 0.00)} & 23.96\textsubscript{(\textpm 0.00)} & 77.94\textsubscript{(\textpm 0.17)} \\ 

& CLIP-KNN & 29.77\textsubscript{(\textpm 0.00)} & 8.64\textsubscript{(\textpm 0.00)} & 95.92\textsubscript{(\textpm 0.09)} & 44.97\textsubscript{(\textpm 0.00)} & 23.89\textsubscript{(\textpm 0.00)} & 79.14\textsubscript{(\textpm 0.17)} & 70.37\textsubscript{(\textpm 0.00)} & 47.78\textsubscript{(\textpm 0.00)} & 92.73\textsubscript{(\textpm 0.10)} & 83.28\textsubscript{(\textpm 0.00)} & 63.57\textsubscript{(\textpm 0.00)} & 72.21\textsubscript{(\textpm 0.11)} \\ \grayline

\multirow{2}{*}{RMDS} & DINOv2 & 6.41\textsubscript{(\textpm 0.00)} & 2.72\textsubscript{(\textpm 0.00)} & 98.13\textsubscript{(\textpm 0.09)} & 20.59\textsubscript{(\textpm 0.00)} & 10.51\textsubscript{(\textpm 0.00)} & 87.62\textsubscript{(\textpm 0.17)} & 33.93\textsubscript{(\textpm 0.00)} & 3.90\textsubscript{(\textpm 0.00)} & 94.95\textsubscript{(\textpm 0.08)} & 64.98\textsubscript{(\textpm 0.00)} & 17.07\textsubscript{(\textpm 0.00)} & 77.94\textsubscript{(\textpm 0.17)} \\ 

& CLIP & 17.33\textsubscript{(\textpm 0.00)} & 8.50\textsubscript{(\textpm 0.00)} & 95.92\textsubscript{(\textpm 0.09)} & 25.61\textsubscript{(\textpm 0.00)} & 4.25\textsubscript{(\textpm 0.00)} & 79.14\textsubscript{(\textpm 0.17)} & 38.51\textsubscript{(\textpm 0.00)} & 8.46\textsubscript{(\textpm 0.00)} & 92.73\textsubscript{(\textpm 0.10)} & 66.41\textsubscript{(\textpm 0.00)} & 30.62\textsubscript{(\textpm 0.00)} & 72.21\textsubscript{(\textpm 0.11)} \\ 

\bottomrule
    \end{tabular}}\par}
\end{table*}

\clearpage
\onecolumn
\section{Code}
\label{sec:code}

In this section, we provide code for RBF layers of an MLRBFN. Additionally, we provide the code for our binary cross-entropy loss function. 

\begin{minted}[
frame=lines,
framesep=2mm,
baselinestretch=1.2,
fontsize=\footnotesize,
linenos
]{python}
import torch
import torch.nn as nn
import torch.nn.functional as F
from torch.nn.modules.lazy import _LazyProtocol

def logsubstractexp(tensor, other):
    a = torch.max(tensor, other)
    return a + ((tensor - a).exp() - (other - a).exp()).log()

def log_bce_loss(log_y_pred, y, num_classes):
    y = F.one_hot(y, num_classes=num_classes).float()
    y_flat = y.ravel()
    log_y_pred_flat = log_y_pred.ravel() - 1e-6
    zero = torch.zeros(log_y_pred_flat.shape).to(y.device)

    not_log_y_pred = logsubstractexp(zero, log_y_pred_flat)
    return torch.mean(-1 * ((y_flat * log_y_pred_flat) + (1 - y_flat) * not_log_y_pred))

def webscale_kmeans(data, centroid_shape, k, iter=1):
    v = torch.zeros(centroid_shape[0]).to(data.device)
    d = torch.zeros(data.shape[0]).int().to(data.device)

    centroids = torch.empty(centroid_shape).to(data.device)
    centroids[0, :] = data[0]
    for c in range(1, centroids.shape[0]):
        dists = torch.cdist(data, centroids[:c], p=k) ** k
        dists = torch.min(dists, -1).values
        dists = dists / torch.sum(dists)

        rand_unif = torch.rand(1).to(data.device)
        dists_cumsum = torch.cumsum(dists, 0)
        idx = torch.argmax((dists_cumsum > rand_unif).to(torch.long))
        centroids[c, :] = data[idx, :] + 1e-5 * torch.randn(data.shape[1]).to(data.device)

    for t in range(iter):
        for i, x in enumerate(data):
            x = x[None, :]
            dists = torch.linalg.norm(centroids - x, ord=k, dim=1) ** k
            idx = torch.argmin(dists)
            d[i] = idx

        for i, x in enumerate(data):
            c = d[i]
            v[c] += 1
            eta = 1 / v[c]
            centroids[c] = (1 - eta) * centroids[c] + eta * x
    
    return centroids

class RBFDepressionLayer(nn.Module):
    def __init__(self, input_features, num_centroids, projection_features, k=2, last=False):
        super().__init__()
        self.centroids = nn.Parameter(torch.randn(num_centroids, input_features))
        self.init_beta = 1
        self.beta = nn.Parameter(torch.ones(num_centroids,))
        self.projections = nn.Parameter(torch.randn(num_centroids, projection_features))
        self.k = k

        self.last = last

    def forward(self, x, depression, recovery=1.1):
        distances = torch.cdist(x, self.centroids, p=self.k) ** self.k
        sb = F.softplus(self.beta)

        if self.last == False:
            scales = (sb / F.softplus(self.init_beta)) * torch.exp(-sb * distances)
            depressed_scales = scales * \
                torch.minimum(depression * recovery, torch.tensor(1.0).to(x.device))[:, None]
            depress_next = torch.max(depressed_scales, 1).values
            x = depressed_scales @ self.projections
            return x, depress_next, scales
        else:
            log_scales = -sb * distances
            log_recovery = torch.log(torch.tensor(recovery).to(x.device))
            log_depression = torch.log(depression)
            x = log_scales + \
                torch.minimum(log_depression + log_recovery, torch.tensor(0.0).to(x.device))[:, None]
            return x, log_depression, log_scales
        
class KMeansRBFDepressionLayer(nn.modules.lazy.LazyModuleMixin, RBFDepressionLayer):
    cls_to_become: RBFDepressionLayer
    centroids: nn.parameter.UninitializedParameter
    beta: nn.parameter.UninitializedParameter

    def __init__(self, input_features, num_centroids, projection_features, k=2, last=False):
        super().__init__(input_features, num_centroids, projection_features, k, last)
        self.centroid_shape = self.centroids.shape
        self.beta_shape = self.beta.shape

        self.centroids = nn.parameter.UninitializedParameter()
        self.beta = nn.parameter.UninitializedParameter()

    def initialize_parameters(self, x, depression, recovery=1.1):
        self.centroids.materialize(self.centroid_shape)
        self.beta.materialize(self.beta_shape)

        x_centroid_loc = x[:x.shape[0]//2, :]
        x_beta = x[x.shape[0]//2:, :]

        cents = webscale_kmeans(x_centroid_loc, self.centroid_shape, self.k, iter=100)
        self.centroids[:] = cents

        distances = torch.cdist(x_beta, self.centroids, p=self.k) ** self.k
        distances_dp = torch.quantile(torch.min(distances, 1).values, 0.95)
        distances_cent = torch.quantile(torch.min(distances, 0).values, 0.95)
        dist = torch.maximum(distances_dp, distances_cent)
        
        if ((4 / dist)) < 5:
            bs = ((4 / dist)).expm1().clamp_min(1e-6).clamp_max(1e4).log()
        else:
            bs = (4 / dist)
        self.init_beta = bs
        self.beta[:] = bs
\end{minted}

\end{document}